\documentclass[10pt,twocolumn,letterpaper]{article}

\usepackage[pagenumbers]{cvpr} 

\usepackage{cuted} 

\usepackage{currfile} 

\usepackage{caption} 

\usepackage[T1]{fontenc}
\usepackage{url}
\usepackage{nicefrac}
\usepackage{placeins}       
\usepackage{multirow}       
\usepackage{algorithm}      
\usepackage{algpseudocode}
\usepackage{colortbl}       
\colorlet{ours}{yellow!30}
\usepackage{pifont}         
\newcommand{\cmark}{\ding{51}}
\newcommand{\xmark}{\ding{55}}

\definecolor{cvprblue}{rgb}{0.21,0.49,0.74}
\usepackage[pagebackref,breaklinks,colorlinks,allcolors=cvprblue]{hyperref}

\usepackage[capitalize]{cleveref}
\crefname{section}{Sec.}{Secs.}
\Crefname{section}{Section}{Sections}
\Crefname{table}{Table}{Tables}
\crefname{table}{Tab.}{Tabs.}

\def\confName{CVPR}
\def\confYear{2027}

\title{VistaCast: Global Subseasonal-to-Seasonal Weather Forecasting with One-Step Generative Ensembles}

\author{Zhewen Hou\\
School of Artificial Intelligence, University of Chinese Academy of Sciences
\and
Chuang Wang\thanks{Corresponding author. Email: \href{mailto:wangchuang@ia.ac.cn}{\texttt{wangchuang@ia.ac.cn}}.}\\
Institute of Automation, Chinese Academy of Sciences\\
School of Artificial Intelligence, University of Chinese Academy of Sciences
}

\hypersetup{
pdftitle={VistaCast: Global Subseasonal-to-Seasonal Weather Forecasting with One-Step Generative Ensembles},
pdfauthor={Zhewen Hou and Chuang Wang},
pdfkeywords={Subseasonal weather forecasting, Generative modeling, AI for Science},
}

\begin{document}
\maketitle

\begin{abstract}
Accurate subseasonal-to-seasonal (S2S) forecasting is critical for resource planning and disaster preparedness in a changing climate.
However, it remains challenging due to the chaotic nature of the atmosphere.
Deterministic AI models suffer from oversmoothing and error accumulation, while generative models are too costly for optimization and inference.
In this paper, we present VistaCast, a global forecasting model that couples a deterministic Spherical Fourier Neural Operator with a one-step generator to produce 44-day ensemble forecasts.
With flow matching, our generator maps source noise to state-dependent forecast residuals in a single network evaluation, making it computationally tractable to finetune multi-member ensembles across long autoregressive trajectories using the continuous ranked probability score (CRPS).
We initialize generation from a physics-informed source distribution that encodes the cross-variable spectral covariance, spatial error variance, and temporal correlation of forecast residuals.
Extensive experiments on the ERA5 reanalysis dataset demonstrate that our model matches or exceeds the forecast skill of traditional numerical methods and leading data-driven models, with only a fraction of the training and inference cost.
Ablation studies show that our ensemble finetuning over long rollouts improves accuracy and that our noise source design better preserves physical consistency.
\end{abstract}


\section{Introduction}
\label{sec:intro}

Subseasonal-to-seasonal (S2S) forecasting predicts atmospheric conditions about two to six weeks in advance~\citep{merryfieldCurrentEmergingDevelopments2020}.
Skillful predictions at this timescale are vital for agricultural planning, energy scheduling, water management, and disaster preparedness~\citep{merryfieldCurrentEmergingDevelopments2020, camps-vallsArtificialIntelligenceModeling2025}.
However, S2S forecasting remains challenging due to the chaotic nature of atmospheric evolution~\citep{merryfieldCurrentEmergingDevelopments2020}.

Over the past few decades, traditional numerical weather prediction (NWP) models that integrate the governing physical equations on supercomputers have steadily advanced forecast skill~\citep{bauerQuietRevolutionNumerical2015, sahaNCEPClimateForecast2014}.
Recently, data-driven weather models have achieved notable advances in medium-range (less than two weeks) weather forecasting~\citep{pathakFourCastNetGlobalDatadriven2022, biAccurateMediumrangeGlobal2023, lamLearningSkillfulMediumrange2023, bonevFourCastNet3Geometric2025, raspWeatherNext3Increasing2026a}.
Trained on high-quality reanalysis data (the observational reconstruction of historical weather), these models have demonstrated superior accuracy with orders-of-magnitude lower inference cost than ECMWF-IFS, the gold standard in operational weather prediction.

However, extending these models to the S2S horizon usually faces severe performance degradation~\citep{nathanielChaosBenchMultiChannelPhysicsBased2024} with two fundamental challenges.
Data-driven deterministic models suffer from oversmoothing at long lead times.
Since models are typically trained with a mean squared error (MSE) objective, they predict the expectation of future states and average across possible trajectories. 
This leads to the blurring of spatial structures and the failure to predict extreme events~\citep{nathanielChaosBenchMultiChannelPhysicsBased2024, langAIFSCRPSEnsembleForecasting2026}.
To make data-driven models truly applicable in operational S2S forecast systems, it is essential to develop probabilistic forecasting models, which generate multiple trajectories (ensemble members) from a given initial state. 
Generative modeling methods, notably diffusion-based models, have demonstrated strong skill in medium-range probabilistic weather prediction~\citep{priceProbabilisticWeatherForecasting2025}. 
Nevertheless, their substantial computational demands, arising from costly training and iterative denoising during inference, potentially limit their reproducibility and broader participation from the academic community~\citep{cachayUCastSurprisinglySimple2026}.

The second challenge is error accumulation over long autoregressive rollouts~\citep{nathanielChaosBenchMultiChannelPhysicsBased2024}.
Since autoregressive forecasters feed each daily prediction back as the next input, the biases and discrepancies keep amplifying as the lead time grows.
It has been shown that, for deterministic models, this can be mitigated through rollout finetuning that backpropagates errors across multiple steps~\citep{lamLearningSkillfulMediumrange2023, guoMaximizingImpactDeep2024}.
However, applying the same strategy to generative models is computationally prohibitive.
When each forecast step takes tens of network evaluations, training over long multi-member rollouts leads to long chains of backpropagation.
This restricts existing generative weather models to training over short horizons and leaves long-range error growth unconstrained~\citep{nguyenOmniCastMaskedLatent2025, couaironArchesWeatherGenSkillfulComputeefficient2026}, which limits  their S2S skill.

To alleviate both problems, we introduce VistaCast, a global subseasonal-to-seasonal ensemble forecasting system that couples a deterministic backbone with a one-step generative residual model.
The deterministic network predicts the mean of evolution trajectories, and the generator samples forecast residuals to restore fine-scale weather structure and represent uncertainty.
Both networks build upon the Spherical Fourier Neural Operator (SFNO)~\citep{bonevSphericalFourierNeural2023}, which operates in the spherical harmonic domain, respecting Earth's spherical geometry~\citep{bonevAttentionSphere2025}.
For residual generation, we adopt improved MeanFlow (iMF)~\citep{gengMeanFlowsOnestep, gengImprovedMeanFlows} to learn the averaged velocity field, mapping source noise to a forecast residual in a single network evaluation.
Critically, single-step sampling makes it computationally tractable to finetune the generator across multi-member trajectories over long autoregressive rollouts.
Furthermore, we start generation from a parameter-free, physics-informed source distribution derived from the empirical error statistics of the deterministic backbone~\citep{derberReformulationBackgroundError1999}, encoding cross-variable spectral covariances, geographic error amplitudes, and time correlations to simplify one-step transport.

We evaluate VistaCast against leading NWP models and state-of-the-art AI models, following ChaosBench~\citep{nathanielChaosBenchMultiChannelPhysicsBased2024} and using the ERA5 reanalysis as ground truth~\citep{hersbachERA5GlobalReanalysis2020}.
In summary, our main contributions are as follows.
\begin{itemize}
	\item \textbf{Enabling long-horizon autoregressive finetuning for generative ensembles.}
	One-step generation makes it computationally feasible to finetune generative ensemble forecasts through full six-week autoregressive rollouts.
	\item \textbf{A physics-informed source distribution.}
	We construct a noise source that encodes the cross-variable, spatial, and temporal structure of empirical forecast errors, simplifying the one-step transport and improving physical consistency.
	\item \textbf{Leading S2S forecast accuracy with high efficiency.}
	VistaCast achieves competitive or superior forecast skill compared with leading NWP and AI models at substantially lower computational cost.
\end{itemize}


\section{Related work}
\label{sec:related}

\begin{figure*}[t]
	\centering
	\includegraphics[width=0.9\linewidth]{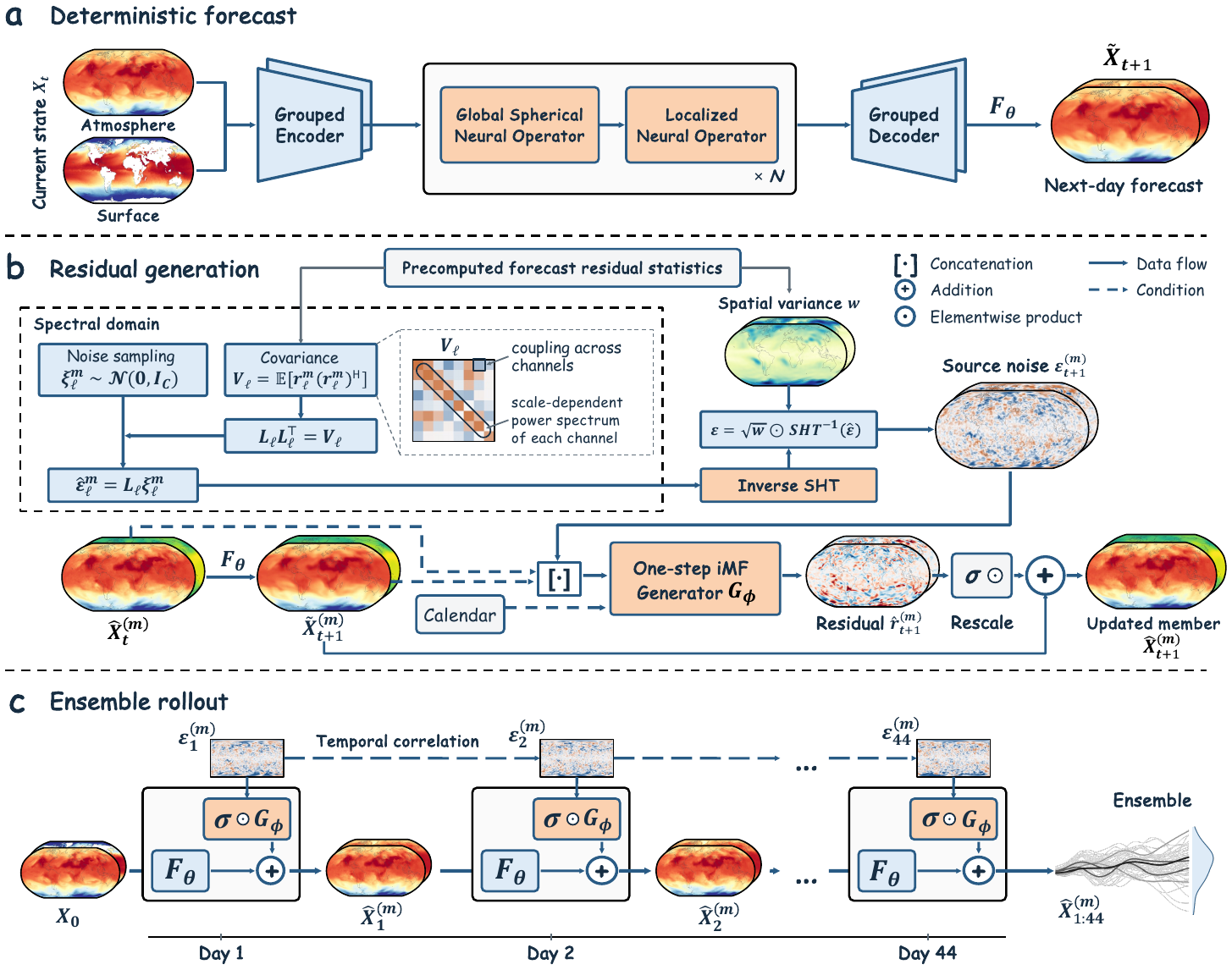}
	\caption{
		Overview of the VistaCast architecture.
		(a) The deterministic backbone $F_\theta$, a spherical neural operator that predicts the next-day state.
		(b) The one-step generator $G_\phi$ maps a sample from the physics-informed noise source to a forecast residual, which is added to the deterministic forecast.
		(c) Each ensemble member is rolled out autoregressively for 44 days with one evaluation of $F_\theta$ and $G_\phi$ per day.
	}
	\label{fig:framework}
\end{figure*}

\subsection{Data-driven weather forecasting}
Deep learning has recently emerged as a powerful alternative to NWP systems for global weather prediction.
Prominent data-driven models such as Pangu-Weather~\citep{biAccurateMediumrangeGlobal2023} and GraphCast~\citep{lamLearningSkillfulMediumrange2023} have achieved notable success in medium-range forecasting, matching or surpassing leading traditional NWP systems at a fraction of the computational cost~\citep{pathakFourCastNetGlobalDatadriven2022, chenFuXiCascadeMachine2023, couaironArchesWeatherGenSkillfulComputeefficient2026}.
To mitigate error accumulation during autoregressive rollouts, these models typically adopt multi-step finetuning, with FengWu~\citep{chenFengWuPushingSkillful2023} and Aurora~\citep{bodnarFoundationModelEarth2025a} further incorporating replay buffers to stabilize long-lead prediction.
Extending these successes to the subseasonal-to-seasonal (S2S) timescale remains difficult due to the chaotic nature of atmospheric dynamics.
ChaosBench~\citep{nathanielChaosBenchMultiChannelPhysicsBased2024}, a comprehensive benchmark for global S2S prediction, reveals that state-of-the-art data-driven models struggle to extend to S2S timescales~\citep{guoMaximizingImpactDeep2024} as error grows.
Although multi-step finetuning helps stabilize long-term forecasts, it becomes computationally prohibitive at S2S scale~\citep{nguyenOmniCastMaskedLatent2025}.
FuXi-S2S~\citep{chenMachineLearningModel2024}, which incorporates a learnable latent perturbation module, is the first data-driven model to reach S2S skill comparable to operational NWP systems.
To sidestep error accumulation, recent data-driven S2S models such as CirT~\citep{liuCirTGlobalSubseasonaltoSeasonal2025a}, TelePiT~\citep{lyuPhysicsInformedTeleconnectionAwareTransformer2025}, and TianQuan-S2S~\citep{liTIANQUANS2SSUBSEASONALTOSEASONALGLOBAL2026} adopt direct prediction strategies and integrate domain-specific priors including spherical patching, teleconnection modeling, and climatological states, though this temporal aggregation sacrifices day-to-day weather evolution.

\subsection{Efficient probabilistic weather forecasting}
Beyond deterministic predictions, recent research has increasingly shifted toward probabilistic modeling to quantify uncertainty and avoid oversmoothing~\citep{priceProbabilisticWeatherForecasting2025, couaironArchesWeatherGenSkillfulComputeefficient2026, raspWeatherNext3Increasing2026a}.
Current methods largely follow two computationally demanding paths.
Generative models based on diffusion or flow matching, such as GenCast~\citep{priceProbabilisticWeatherForecasting2025}, produce detailed spatial structures, yet their reliance on iterative denoising steps introduces substantial inference cost.
In contrast, models directly optimized with CRPS, such as FourCastNet~3~\citep{bonevFourCastNet3Geometric2025} and AIFS-CRPS~\citep{langAIFSCRPSEnsembleForecasting2026}, enable relatively fast inference, yet impose heavy training costs since they must generate a full ensemble at each update to evaluate the loss.

Several recent works seek to improve efficiency in medium-range probabilistic forecasting.
U-Cast~\citep{cachayUCastSurprisinglySimple2026} streamlines both training and inference by combining a compact backbone with simple Monte Carlo dropout for stochasticity.
Tyche~\citep{xuTycheOneStep2026} also accelerates generative sampling through one-step flow transport, yet we argue that mapping raw noise directly to full atmospheric states in a single step poses optimization difficulties given the complexity of weather evolution.
Moreover, existing generative models are restricted to short-horizon autoregressive finetuning under two weeks.
At the S2S scale, we argue that long rollout training is more critical since S2S forecasting prioritizes stable prediction beyond two weeks over short-lead precision.
To the best of our knowledge, VistaCast is the first generative forecaster to bridge this gap.
By decoupling the deterministic trajectory mean from stochastic residuals and pairing a generator with a physics-informed source distribution, our approach simplifies the one-step transport and allows a lightweight generator to support both efficient long-horizon training and fast inference.


\section{Preliminaries}
\label{sec:preliminaries}

We formulate subseasonal forecasting as sampling an ensemble of daily global atmospheric trajectories up to 44 days ahead.
Let $X_t \in \mathbb{R}^{C \times H \times W}$ denote the atmospheric state on day $t$, comprising $C = 59$ channels (five upper-air variables across ten vertical pressure levels and nine surface variables) on a $1.5^\circ$ latitude--longitude grid of $H \times W = 121 \times 240$ points.
Given an initial state $X_{t_0}$, our goal is to sample $M$ trajectories $\{\hat{X}^{(m)}_{t_0+1:t_0+T}\}_{m=1}^{M}$ over $T = 44$ days.

VistaCast decomposes each daily transition into a deterministic prediction and a generated forecast residual (Figure~\ref{fig:framework}) with
\begin{equation}
	\begin{gathered}
	\tilde{X}^{(m)}_{t+1} = F_\theta\big(\hat{X}^{(m)}_t\big), \\
	\hat{X}^{(m)}_{t+1} = \tilde{X}^{(m)}_{t+1} + \sigma \odot G_\phi\big(\varepsilon^{(m)}_{t+1};\, \hat{X}^{(m)}_t, \tilde{X}^{(m)}_{t+1}\big),
	\end{gathered}
	\label{eq:transition}
\end{equation}
where $\hat{X}^{(m)}_{t_0} = X_{t_0}$, $F_\theta$ is a deterministic SFNO that predicts the daily evolution, $G_\phi$ is a conditional flow generator that maps a noise sample $\varepsilon^{(m)}_{t+1}$ to a standardized forecast residual, $\sigma \in \mathbb{R}^{C}$ rescales the residual, and $\odot$ denotes channelwise multiplication.

\paragraph{Spherical Fourier neural operators.}
Neural operators learn mappings between continuous function spaces~\citep{kovachkiNeuralOperatorLearning2023}.
The Fourier neural operator (FNO) parameterizes global kernel integrals as learnable multiplications in Fourier space~\citep{liFourierNeuralOperator2021}.
To incorporate the spherical geometry of the Earth, the Spherical Fourier Neural Operator (SFNO)~\citep{bonevSphericalFourierNeural2023} operates in the basis of spherical harmonics via the spherical harmonic transform (SHT),
\(f(\vartheta, \lambda) = \sum_{\ell=0}^{L} \sum_{m=-\ell}^{\ell} \hat{f}_\ell^m\, Y_\ell^m(\vartheta, \lambda),\)
where $\vartheta$ and $\lambda$ denote colatitude and longitude, $Y_\ell^m$ are the spherical harmonics, and the total degree $\ell$ indexes the horizontal spatial scale.
Each SFNO block mixes feature channels at every degree through $\hat{z}_\ell^m = W_\ell\, \hat{v}_\ell^m$ with weights shared across zonal orders $m$.
A single SFNO layer provides a global receptive field since each spectral coefficient aggregates information across the entire globe.

\paragraph{Diffusion and flow matching.}
Diffusion models generate samples by learning to reverse a gradual noising process.
Let $r$ denote the target and $\varepsilon \sim p_0$ a sample from the source distribution.
Flow matching~\citep{lipmanFlowMatchingGenerative2023} constructs a linear probability path interpolating between the target and source via
\begin{equation}
	z_s = (1 - s)\, r + s\, \varepsilon, \qquad s \in [0, 1],
	\label{eq:path}
\end{equation}
with constant conditional velocity $v = \mathrm{d}z_s / \mathrm{d}s = \varepsilon - r$.
Training a neural network $v_\phi(z_s, s)$ to regress $\varepsilon - r$ recovers the marginal velocity field that transports $p_0$ to the target residual distribution.
Critically, the flow-matching formulation only requires sampling from $p_0$ and does not restrict it to an isotropic Gaussian.
We exploit this property to incorporate physical error priors.


\section{Methodology}
\label{sec:method}

\subsection{Model architecture}
\label{sec:architecture}

\paragraph{Backbone.}
The deterministic model $F_\theta$ and the generator $G_\phi$ share the same encoder--processor--decoder backbone.
An encoder projects input fields onto a latent spherical grid, a processor transforms the latent representations, and a decoder projects them back to the $C$ output channels.
The processor includes global SFNO blocks and localized discrete--continuous spherical convolution (DISCO) blocks, which operate on compact geodesic neighborhoods to preserve sharp local gradients~\citep{liu-schiaffiniNeuralOperatorsLocalized2024}.
$F_\theta$ is conditioned~\citep{perezFiLMVisualReasoning2018} on the target day of year and forecast lead time.
$G_\phi$ takes the concatenation of the intermediate flow state $z_s$, the current state $X_t$, and the deterministic prediction $\tilde{X}_{t+1}$, conditioned on the sum of separate embeddings for the two flow time variables, the day of year, and the forecast lead time.

\paragraph{One-step generation.}
We adopt improved MeanFlow (iMF)~\citep{gengMeanFlowsOnestep, gengImprovedMeanFlows}, which parameterizes the average velocity over a time interval $[s', s]$ via
\begin{equation}
	u(z_s, s', s) = \frac{1}{s - s'} \int_{s'}^{s} v(z_\varsigma, \varsigma)\, \mathrm{d}\varsigma.
	\label{eq:meanflow}
\end{equation}
Evaluating the learned network $u_\phi$ over the full interval $s = 1$ and $s' = 0$ maps a source sample directly to a forecast residual in a single forward pass, $\hat{r} = \varepsilon - u_\phi(\varepsilon, 0, 1)$.
Although the interval-averaged velocity $u$ has no direct analytical target, differentiating Eq.~\eqref{eq:meanflow} yields the consistency identity $v(z_s, s) = u(z_s, s', s) + (s - s')\, \tfrac{\mathrm{d}}{\mathrm{d}s} u(z_s, s', s)$.
Following iMF, we form the composite velocity prediction
\begin{equation}
    V_\phi(z_s,s',s)=u_\phi(z_s,s',s)
    +(s-s')\,\mathrm{sg}[\tfrac{\mathrm{d}}{\mathrm{d}s}u_\phi(z_s,s',s)],
    \label{eq:imf}
\end{equation}
and train $V_\phi$ to regress the conditional velocity $v = \varepsilon - r$, where $\mathrm{sg}[\cdot]$ denotes the stop-gradient operator.
The total derivative $\tfrac{\mathrm{d}}{\mathrm{d}s} u_\phi = u_\phi(z_s, s, s)\, \partial_z u_\phi + \partial_s u_\phi$ evaluates the directional derivative along the model's own instantaneous velocity $u_\phi(z_s, s, s)$~\citep{gengImprovedMeanFlows}.
This Jacobian--vector product is computed efficiently via forward-mode automatic differentiation within the same forward pass that evaluates $u_\phi$.
We optimize the adaptive latitude-weighted objective
\begin{equation}
	\mathcal{L}_{\mathrm{iMF}} = \mathbb{E}_{r, \varepsilon, s', s}\Big[\, w \, \big\| V_\phi(z_s, s', s) - (\varepsilon - r) \big\|_{A}^{2} \Big], 
	\label{eq:imf_loss}
\end{equation}
where $\|\cdot\|_A^2$ denotes the latitude-weighted mean squared norm and $w$ is an adaptive loss weight (Supplementary Section~\ref{sec:supp_imf}).

\subsection{Physics-informed noise source}
\label{sec:noise}

Most generative weather models initialize sampling from isotropic Gaussian white noise drawn independently across grid points, channels, and time steps.
Atmospheric conditions, however, exhibit multi-scale and multivariate correlations.
Requiring a one-step generator to synthesize all of the physical structure of weather states from uncorrelated white noise in a single forward pass places an unnecessary burden on the network.
Drawing inspiration from background-error covariance modeling in variational data assimilation~\citep{derberReformulationBackgroundError1999}, we construct the prior $p_0$ as a structured field whose spatial and cross-variable second-order statistics match the empirical forecast residuals of $F_\theta$ over the training period.

\paragraph{Spectral and inter-variable covariance.}
To capture multivariate physical relationships, we transform the standardized forecast residuals of $F_\theta$ into the spherical harmonic domain and model cross-channel correlations at each spatial scale.
Specifically, we sample the complex spectral coefficients of all $C$ channels via
\begin{gather}
	V_\ell = \frac{1}{N(2\ell+1)} \sum_{t=1}^{N} \sum_{m=-\ell}^{\ell} \mathrm{Re}\Big[\mathrm{SHT}(r_t)_\ell^m \big(\mathrm{SHT}(r_t)_\ell^m\big)^{\mathrm{H}}\Big], \notag \\
	L_\ell L_\ell^\top = V_\ell, \quad \hat{\varepsilon}_\ell^m = L_\ell\, \xi_\ell^m, \quad \xi_\ell^m \sim \mathcal{N}(0, I_C), \label{eq:spectral_noise}
\end{gather}
where $V_\ell$ is the empirical cross-channel covariance at degree $\ell$, averaged over zonal orders $m$ and training days $t$, and $L_\ell$ is its Cholesky factor.
The diagonal entries of $V_\ell$ specify the error power spectrum of each channel as a function of horizontal scale $\ell$.
This concentrates source variance at the planetary and synoptic wavelengths and prevents injecting unphysical grid-scale noise.
The off-diagonal entries of $V_\ell$ encode scale-dependent multivariate error couplings across variables and vertical levels.

\paragraph{Geographic modulation.}
The inverse SHT of $\hat{\varepsilon}$ has the same statistics at every location on the sphere, but actual forecast errors vary in amplitude across regions.
We therefore modulate each channel in grid space according to its empirical variance $w_c(\vartheta, \lambda)$,
\begin{equation}
	\varepsilon_c(\vartheta, \lambda) = \sqrt{w_c(\vartheta, \lambda)}\, \big[\mathrm{SHT}^{-1}\hat{\varepsilon}\big]_c(\vartheta, \lambda).
	\label{eq:varmap}
\end{equation}

\paragraph{Temporal persistence.}
To prevent consecutive daily perturbations from canceling and offsetting, the standard normal draws $\xi_t$ at each degree $\ell$ (with order index $m$ omitted) combine a static component $\xi^{\mathrm{sh}}$ with an AR(1) process,
\begin{equation}
	\begin{gathered}
	\xi_t = \sqrt{\beta_\ell}\, \xi^{\mathrm{sh}} + \sqrt{1 - \beta_\ell}\, \eta_t, \\
	\eta_t = \rho_\ell\, \eta_{t-1} + \sqrt{1 - \rho_\ell^2}\, \zeta_t, 
	\end{gathered}
	\label{eq:ar1}
\end{equation}
where \(\xi^{\mathrm{sh}}, \eta_{t_0}, \zeta_t \sim \mathcal{N}(0, I_C),\)
inducing a lag-$n$ temporal correlation of $\beta_\ell + (1 - \beta_\ell)\, \rho_\ell^{\,n}$.
Moreover, the marginal distribution at every step remains $\xi_t \sim \mathcal{N}(0, I_C)$.

\subsection{Curriculum training and inference}

\begin{figure*}[t]
	\centering
	\includegraphics[width=\linewidth]{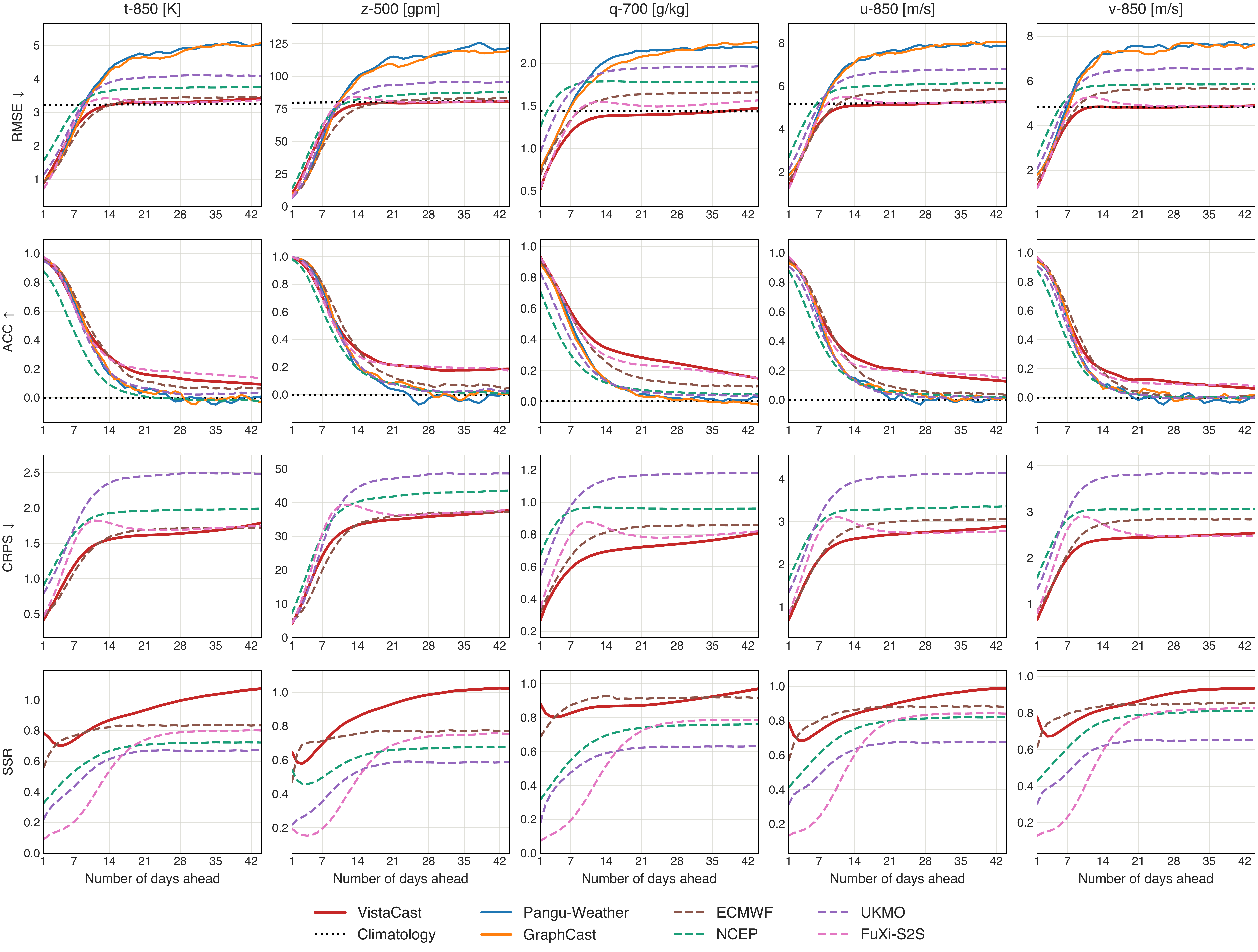}
	\caption{
		Forecast skill on the 2022 test year against lead time for five key variables.
		Rows show (1) deterministic metrics, including RMSE (lower is better) and ACC (higher is better), and (2) probabilistic metrics, including CRPS (lower is better) and spread--skill ratio (ideally 1.0), from top to bottom.
		For deterministic evaluation, VistaCast is represented by its deterministic backbone $F_\theta$, which efficiently approximates the mean trajectory.
		Dashed curves denote probabilistic baselines, and the dotted curve denotes the climatology.
	}
	\label{fig:skill}
\end{figure*}

\paragraph{Deterministic training.}
We train VistaCast in three stages.
Firstly, we train the deterministic backbone $F_\theta$ to predict the mean daily evolution using a rollout curriculum.
Unrolling $F_\theta$ for $K$ daily steps from an initial state, we minimize the accumulated latitude-weighted mean squared error.
Increasing the lead from $K = 1$ to $K = 42$ across curriculum phases exposes $F_\theta$ to its own multi-week forecast states and mitigates error accumulation~\citep{lamLearningSkillfulMediumrange2023, chenMachineLearningModel2024}.
After that, $F_\theta$ is used to calculate the forecast residuals and serves as the control member.

\paragraph{Generative pretraining.}
We then pretrain the one-step generator $G_\phi$ on single-day residual transitions.
We optimize $G_\phi$ with the iMF loss (Eq.~\eqref{eq:imf_loss}) to model the conditional residual distribution $p\big(r_{t+1} \mid X_t, F_\theta(X_t)\big)$ using independent spatial draws from the physics-informed source (Eqs.~\eqref{eq:spectral_noise}--\eqref{eq:varmap}).
To bridge the distribution shift between reanalysis inputs and model-generated rollout states, we subsequently continue single-step training on transitions initialized from short deterministic rollouts $F_\theta^{(k-1)}(X_t)$ with rollout depth $k \in \{1, 2, 3, 4\}$ sampled uniformly.

\paragraph{Ensemble rollout finetuning.}
Autoregressively unrolling a prediction introduces errors that compound across steps.
The deterministic operator $F_\theta$ contracts the ensemble spread and accumulates drift.
In Stage~3, we therefore finetune the coupled ensemble over long rollouts.
From each initial state $X_{t_0}$, we unroll $M = 3$ ensemble members over $H$ daily rollout steps.
We minimize the trajectory-averaged almost-fair CRPS (afCRPS)~\citep{langAIFSCRPSEnsembleForecasting2026},
\begin{equation}
\begin{aligned}
	\mathcal{L}_{\mathrm{roll}}
	&= \frac{1}{H}\sum_{k=1}^{H}\Bigg[
	\frac{1}{M}\sum_{m=1}^{M}
	\big|\hat{X}^{(m)}_{t_0+k}-X_{t_0+k}\big| \\
	&\quad - \frac{1-(1-\alpha)/M}{2M(M-1)}
	\sum_{m=1}^{M}\sum_{n\ne m}
	\big|\hat{X}^{(m)}_{t_0+k}-\hat{X}^{(n)}_{t_0+k}\big|
	\Bigg],
\end{aligned}
\label{eq:afcrps}
\end{equation}
evaluated pointwise with $\alpha = 0.95$ and averaged with latitude weights.
The first term penalizes member error relative to ground truth and the second term rewards inter-member dispersion~\citep{langAIFSCRPSEnsembleForecasting2026}.
Since iMF generates each daily residual in a single evaluation, a 42-day, three-member rollout requires only 126 generator evaluations, compared with thousands for iterative samplers, making long-chain backpropagation computationally feasible.
We progressively extend the rollout step $H$ from 2 to 42 days, and from $H = 10$ on also finetune $F_\theta$ jointly with $G_\phi$.

\paragraph{Ensemble inference.}
At inference, we generate a 51-member ensemble comprising 50 perturbed and one control member.
To calibrate the ensemble spread at long leads, we also apply a mild multiplicative covariance inflation~\citep{andersonMonteCarloImplementation1999} prior to adding each daily residual,
\begin{equation}
	\tilde{X}^{(m)}_{t+1} \leftarrow \tilde{X}^{(m)}_{t+1} + (g_k - 1)\, \mathcal{P}_{10}\big(\tilde{X}^{(m)}_{t+1} - \overline{\tilde{X}}_{t+1}\big),
	\label{eq:inflation}
\end{equation}
where $\overline{\tilde{X}}_{t+1}$ is the ensemble mean of the backbone forecasts, $\mathcal{P}_{10}$ keeps only the large-scale components with spherical harmonic degrees $\ell \le 10$ (acting as a low-pass filter), and the atmospheric gain $g_k$ increases linearly from $1.0$ at initialization to $1.05$ at day 44.


\section{Experiments}
\label{sec:experiments}

We evaluate VistaCast against operational numerical ensembles and leading AI weather models on ChaosBench (Section~\ref{sec:skill}), benchmark its computational efficiency (Section~\ref{sec:efficiency}), and ablate its physics-informed source distribution and trajectory-level rollout finetuning (Section~\ref{sec:ablation}).

\subsection{Experimental setup}
\label{sec:setup}

\paragraph{Dataset.}
We train and evaluate on the $1.5^\circ$ ERA5 reanalysis~\citep{hersbachERA5GlobalReanalysis2020, raspWeatherBench2Benchmark2024a}.
Following ChaosBench, a widely adopted S2S benchmark~\citep{nathanielChaosBenchMultiChannelPhysicsBased2024}, we use 1979--2019 for training, 2020--2021 for validation, and 2022 for testing.
We show five core variables, including 850\,hPa temperature (t-850), 500\,hPa geopotential (z-500), 700\,hPa specific humidity (q-700), and 850\,hPa zonal and meridional winds (u-850 and v-850).

\paragraph{Baselines.}
We compare VistaCast against the leading NWP and data-driven models: (1) Operational NWP models, including the ECMWF IFS extended-range ensemble~\citep{molteniECMWFEnsemblePrediction1996, vitartEvolutionECMWFSubseasonal2014}, NCEP CFSv2~\citep{sahaNCEPClimateForecast2014}, and UKMO GloSea6~\citep{kettleboroughGlobalSeasonalForecasting2025}.
(2) Medium-range AI models, including Pangu-Weather~\citep{biAccurateMediumrangeGlobal2023} and GraphCast~\citep{lamLearningSkillfulMediumrange2023}, rolled out autoregressively to 44 days.
(3) FuXi-S2S~\citep{chenMachineLearningModel2024}, a state-of-the-art AI S2S model.
(4) The daily climatology over 1981--2010 is also added for reference.
Scores for (1) and (2) are obtained from ChaosBench.
FuXi-S2S is evaluated using our own runs, and (4) is calculated with our own data.
We exclude CirT, TelePiT, and TianQuan-S2S from the main comparison due to mismatched prediction targets, the limited comparability of deterministic biweekly-mean forecasts with daily ensembles, and the lack of publicly available code for TianQuan-S2S.
Additional comparisons with CirT and TelePiT are provided in Supplementary Section~\ref{sec:supp_biweekly}.

\paragraph{Evaluation metrics.}
Deterministic skill is evaluated using root mean square error (RMSE) and anomaly correlation coefficient (ACC). 
Probabilistic skill is evaluated with the continuous ranked probability score (CRPS)~\citep{gneitingStrictlyProperScoring2007}.
Calibration is measured by the spread--skill ratio (SSR) with an ideal value of 1.0.
All scores are latitude-weighted and averaged over all initialization dates.

\paragraph{Implementation details.}
$F_\theta$ (30.6\,M parameters) and $G_\phi$ (16.4\,M) are trained with AdamW~\citep{loshchilovDecoupledWeightDecay2019} (weight decay 0.01) in bf16-mixed precision on NVIDIA A800 GPUs.
Stages~1 and~3 lengthen the rollout to 42 days over 15 and 13 curriculum phases with decreasing batch size and learning rate, while Stage~2 trains $G_\phi$ for 37k iterations on single-day transitions.
The full pipeline takes under 6 A800 GPU-days.
Complete hyperparameters (especially for the training curriculum) are listed in Supplementary Section~\ref{sec:supp_training}.

\begin{table}[t]
	\caption{
		Ablations of VistaCast over sampling steps and source components.
		$\Delta$CRPS (relative to the highlighted full model) and SSR are averaged over weeks 3--4 and 5--6 and summarized by the median over all 59 variables.
		Time is per member and forecast day on one A800 GPU, averaged over all variants.
		Color denotes the spectral covariance and spatial variance of the source, and Temp.\ its temporal persistence.
		Bold marks the best value within each block.
	}
	\label{tab:ablation}
	\centering
	\footnotesize
	\setlength{\tabcolsep}{3pt}
	\setlength{\aboverulesep}{0pt}
	\setlength{\belowrulesep}{0pt}
	\renewcommand{\arraystretch}{1.1}
	\newcommand{\headstrut}{\rule[-0.375\baselineskip]{0pt}{1.25\baselineskip}}  
	\begin{tabular}{c|c|cc|*{4}{w{c}{23.7pt}}|c}
		\toprule
		\headstrut& \multirow{2}{*}{\textbf{NFE}} & \multicolumn{2}{c|}{\textbf{Source}} & \multicolumn{2}{c}{\textbf{\boldmath$\Delta$CRPS (\%) $\downarrow$}} & \multicolumn{2}{c|}{\textbf{\boldmath SSR $\rightarrow 1$}} & \textbf{Time} \\
		\cmidrule(lr){3-4} \cmidrule(lr){5-6} \cmidrule(lr){7-8}
		\headstrut& & Color & Temp. & W3-4 & W5-6 & W3-4 & W5-6 & (ms) \\
		\midrule
		\multirow{6}{*}{\rotatebox[origin=c]{90}{\begin{tabular}{@{}c@{}}\textbf{Sampling steps}\end{tabular}}} & \multirow{2}{*}{1} & \cellcolor{ours}\cmark & \cellcolor{ours}\cmark & \cellcolor{ours}\textbf{0.0} & \cellcolor{ours}\textbf{0.0} & \cellcolor{ours}\textbf{0.93} & \cellcolor{ours}\textbf{0.99} & \multirow{2}{*}{47} \\
		 &  & \xmark & \xmark & +2.0 & +6.0 & 0.92 & 0.97 &  \\
		\cmidrule{2-9}
		 & \multirow{2}{*}{2} & \cmark & \cmark & +0.5 & +1.4 & 0.89 & 1.03 & \multirow{2}{*}{66} \\
		 &  & \xmark & \xmark & +3.9 & +9.0 & 0.83 & 0.94 &  \\
		\cmidrule{2-9}
		 & \multirow{2}{*}{4} & \cmark & \cmark & +1.3 & +3.0 & 0.87 & 1.05 & \multirow{2}{*}{103} \\
		 &  & \xmark & \xmark & +8.4 & +13.0 & 0.70 & 0.83 &  \\
		\midrule
		\multirow{4}{*}{\rotatebox[origin=c]{90}{\begin{tabular}{@{}c@{}}\textbf{Source}\end{tabular}}} & \multirow{4}{*}{1} & \cellcolor{ours}\cmark & \cellcolor{ours}\cmark & \cellcolor{ours}\textbf{0.0} & \cellcolor{ours}\textbf{0.0} & \cellcolor{ours}0.93 & \cellcolor{ours}\textbf{0.99} &  \\
		 &  & \cmark & \xmark & +0.6 & +0.2 & 0.83 & 0.84 &  \\
		 &  & \xmark & \cmark & +1.0 & +3.9 & \textbf{0.98} & 0.98 &  \\
		 &  & \xmark & \xmark & +2.0 & +6.0 & 0.92 & 0.97 &  \\
		\bottomrule
	\end{tabular}
\end{table}

\begin{figure}[t]
	\centering
	\includegraphics[width=\linewidth]{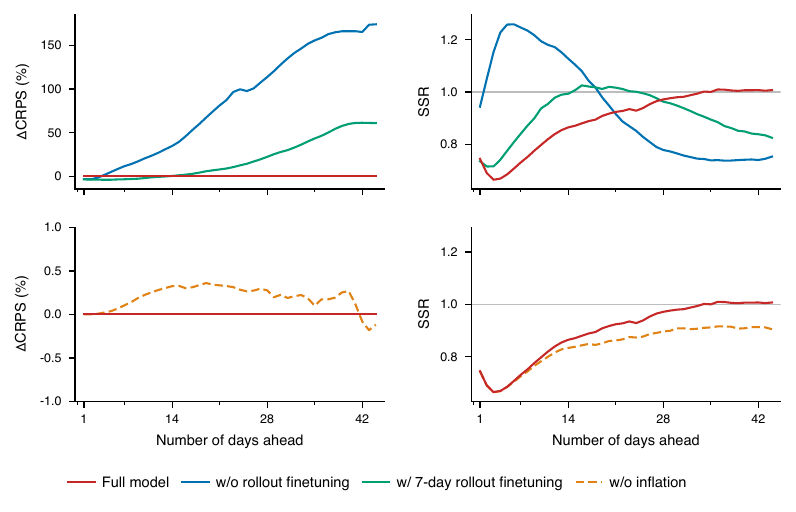}
	\caption{
		Effect of rollout finetuning (top) and covariance inflation (bottom).
		The full model is VistaCast with 42-day rollout finetuning and inflation.
		Left: CRPS change relative to the full model, as the median over all 59 variables.
		Right: median SSR.
	}
	\label{fig:ablation_curves}
\end{figure}

\subsection{Forecast skill}
\label{sec:skill}

\paragraph{Deterministic skill.}
We compare the deterministic skill of VistaCast against the baselines with RMSE and ACC (top two rows of Figure~\ref{fig:skill}).
Although Pangu-Weather and GraphCast are accurate in the first week, their autoregressive trajectories diverge rapidly beyond two weeks, and their forecast errors soon exceed those of the climatology.
Our model is on par with the performance of medium-range models in weeks~1 and~2 and remains stable across all lead times.
Over the entire period, our model matches or exceeds the forecast skill of traditional numerical methods and leading data-driven S2S models.

\paragraph{Probabilistic skill.}
The bottom two rows of Figure~\ref{fig:skill} evaluate the probabilistic skill with CRPS and SSR.
Compared with FuXi-S2S, our model achieves comparable or lower CRPS on multiple variables, especially before 21 days of lead time.
Against operational NWP ensembles, VistaCast consistently outperforms NCEP and UKMO throughout the S2S window and matches the ECMWF ensemble.
Beyond the second week, VistaCast achieves the best calibration among all models, with an SSR closest to 1.0 for most variables across the remaining leads.

\subsection{Computational efficiency}
\label{sec:efficiency}

Beyond strong forecast skill, VistaCast offers substantial efficiency gains over existing methods.
The complete training curriculum takes under 6 A800 GPU-days, and a 44-day forecast costs only about 2.1\,s per ensemble member on a single A800 GPU.
FuXi-S2S, a leading probabilistic S2S prediction model at the same 1.5$^\circ$ resolution, is the most comparable work.
Compared with FuXi-S2S, VistaCast uses 22 times fewer parameters (47.0\,M \vs 1.04\,B) and about 50 times fewer TFLOPs per member and step (0.31 \vs 15.1).

At each daily forecast step, our model calls the generator only once, and the physics-informed source distribution relieves the burden on the generator at negligible computational cost.
Beyond accelerating inference, the high computational efficiency brought by this one-step strategy is what makes long rollout finetuning feasible in practice.
If each daily transition required iterative sampling for dozens of steps, the computation graph would grow multiplicatively across steps and quickly exceed GPU memory limits.

\subsection{Ablation studies}
\label{sec:ablation}

\begin{figure*}[t]
	\centering
	\includegraphics[width=\linewidth]{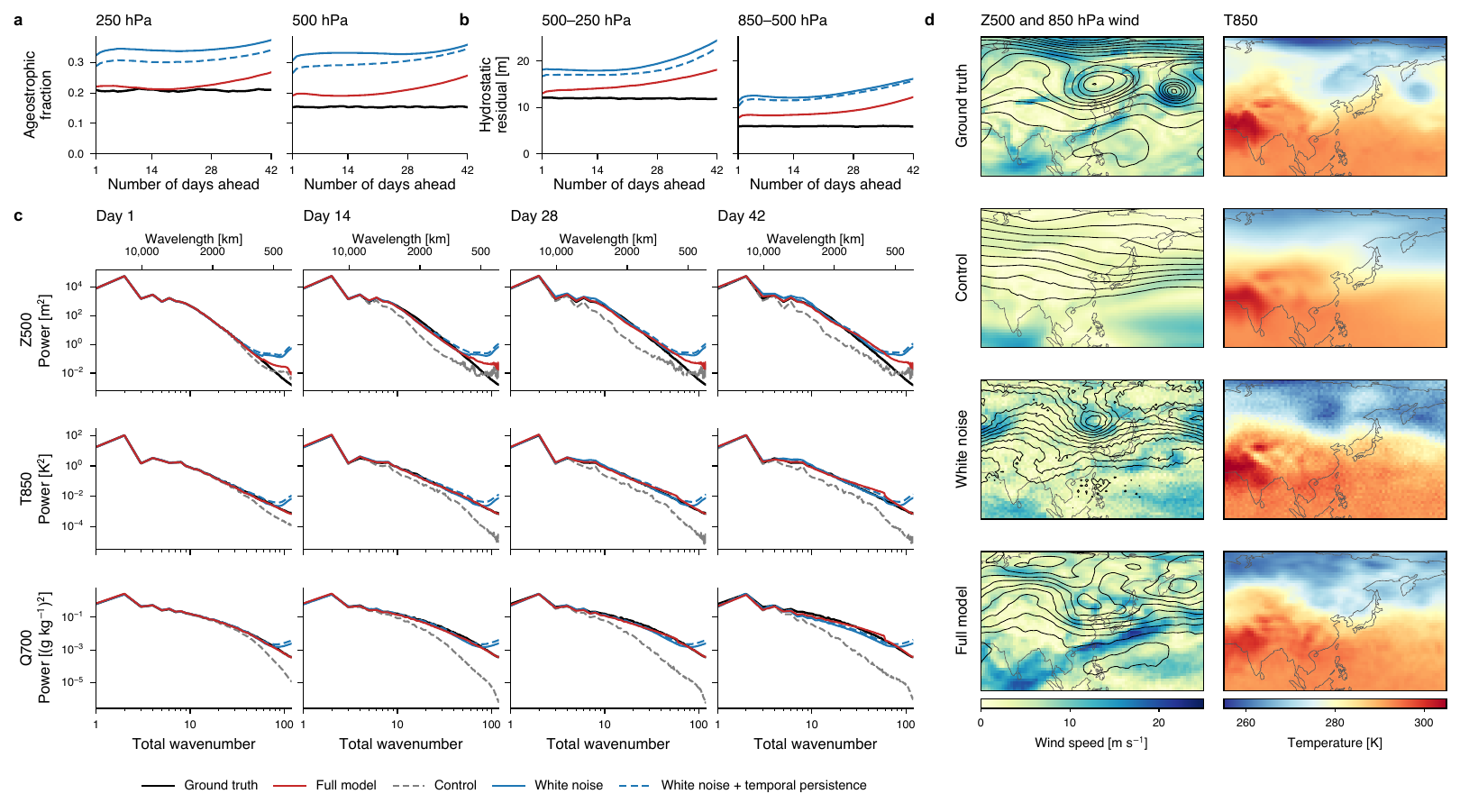}
	\caption{
		Spatial structure and physical consistency ablation of individual ensemble members.
		(a) Ageostrophic wind fraction at 250 and 500\,hPa averaged over 20--70$^\circ$ latitude and (b) hydrostatic residual of the 500--250 and 850--500\,hPa layers.
		(c) Spherical harmonic power spectra of Z500, T850, and Q700 at days 1, 14, 28, and 42.
		(d) Example of day-28 maps of Z500 (contours) over 850\,hPa wind speed (shading) and T850 initialized on 14 April 2022.
		Control is the deterministic model.
		White noise replaces our physics-informed source with isotropic Gaussian noise, and White noise + temporal persistence also adds temporal persistence to the source.
	}
	\label{fig:structure}
\end{figure*}

We ablate the key components of VistaCast while keeping all other settings fixed during training and inference.
Results are shown in Table~\ref{tab:ablation} and Figure~\ref{fig:ablation_curves}. 

Surprisingly, one-step sampling (NFE = 1) achieves the best forecast skill with the fastest inference speed for both noise sources, whereas increasing the sampling steps to NFE~2 and NFE~4 slightly degrades performance. 
Although somewhat counterintuitive, this behavior can be attributed to the rollout finetuning under one-step generation, which makes training and inference most consistent at NFE~1.
This further suggests that single-step generation and rollout finetuning could complement each other.
We next isolate the contributions of the noise components, namely coloring (imposing spectral covariance and spatially varying variance on the source noise) and temporal persistence.  
In the source part of Table~\ref{tab:ablation}, spatial coloring yields modest CRPS gains and trails slightly in SSR.
We would argue later that the role of spatial coloring lies in improving physical consistency and spatial realism, which pointwise CRPS cannot capture. 

Meanwhile, temporal persistence markedly improves ensemble calibration when combined with colored noise.
This is likely because independent daily perturbations tend to cancel out across consecutive steps.
Rollout finetuning has a notably large impact, bringing substantial improvements across weeks~3 to~6.
Finetuning beyond day~7 also significantly improves CRPS.
This indicates that long-lead skill gains benefit greatly from training over extended horizons. 
Covariance inflation provides a mild calibration adjustment that leaves CRPS nearly unchanged and shifts SSR toward a more reasonable range.

To evaluate physical consistency, we examine geostrophic and hydrostatic balance and spherical harmonic power spectra (Figure~\ref{fig:structure}). 
In terms of spatial structure, predictions from the deterministic model become visibly blurred, with small-scale spectral power dropping below that of ERA5.
The white-noise variant introduces too much grid-scale noise and causes an unphysical upward tilt in the power spectrum. 
The full model tracks the ERA5 spectrum more closely and produces more realistic synoptic patterns in the forecast maps. 
For the tested physical balances, VistaCast also stays much closer to ERA5 than the white-noise variant.
Because the generator is conditioned on lead time to adapt its residuals to growing forecast uncertainty, this physical deviation grows gradually as the forecast lead time increases.
In addition, adding only temporal persistence to white noise brings little to no improvement.

Overall, these results validate our noise design, where spectral covariance and spatial variance preserve physical consistency while temporal persistence calibrates ensemble spread.


\FloatBarrier
\section{Discussion and conclusion}
\label{sec:conclusion}

We have presented VistaCast, a global S2S ensemble forecasting model that uses a deterministic SFNO backbone to predict the mean atmospheric evolution and a one-step improved MeanFlow (iMF) generator to model forecast residuals.
On ChaosBench, VistaCast matches or surpasses operational NWP ensembles and leading AI models at a small fraction of their computational cost, while producing ensemble members that retain realistic synoptic structures and multivariate balance.

The main lesson of our work is that pairing one-step flow matching with a physically structured noise source achieves efficient generation with high quality.
Similar approaches could be useful beyond weather forecasting.
When few-step flow matching is applied, embedding domain-specific inductive biases directly into the source distribution could offer a general route to making the transport from noise to data less arduous.
Future work could pursue scaling of model capacity and training data and coupling with slower Earth-system components such as the ocean, land surface, and sea ice.

\section*{LLM Usage Statement}
Large language models, specifically GPT-5.6 Sol, Claude Opus 5 and 5.5, and Gemini 4 Argon, were employed in preparing this paper.
They assisted with refining grammar and improving the clarity, coherence, and technical readability of the manuscript.
LLMs were widely used for code development, review, and debugging under the authors' direction.
They were sparingly used for research idea articulation and experimental analysis.
No LLMs were used for data generation.
We carefully inspected and verified all LLM-assisted text and code, and we take full responsibility for all content of the paper.

{
\small
\setlength{\emergencystretch}{3em}
\bibliographystyle{ieeenat_fullname}
\bibliography{references}

\begin{thebibliography}{42}
\providecommand{\natexlab}[1]{#1}
\providecommand{\url}[1]{\texttt{#1}}
\expandafter\ifx\csname urlstyle\endcsname\relax
  \providecommand{\doi}[1]{doi: #1}\else
  \providecommand{\doi}{doi: \begingroup \urlstyle{rm}\Url}\fi

\bibitem[Anderson and Anderson(1999)]{andersonMonteCarloImplementation1999}
Jeffrey~L. Anderson and Stephen~L. Anderson.
\newblock A {Monte Carlo} implementation of the nonlinear filtering problem to
  produce ensemble assimilations and forecasts.
\newblock \emph{Monthly Weather Review}, 127\penalty0 (12):\penalty0
  2741--2758, 1999.

\bibitem[Bauer et~al.(2015)Bauer, Thorpe, and
  Brunet]{bauerQuietRevolutionNumerical2015}
Peter Bauer, Alan Thorpe, and Gilbert Brunet.
\newblock The quiet revolution of numerical weather prediction.
\newblock \emph{Nature}, 525\penalty0 (7567):\penalty0 47--55, 2015.

\bibitem[Bi et~al.(2023)Bi, Xie, Zhang, Chen, Gu, and
  Tian]{biAccurateMediumrangeGlobal2023}
Kaifeng Bi, Lingxi Xie, Hengheng Zhang, Xin Chen, Xiaotao Gu, and Qi Tian.
\newblock Accurate medium-range global weather forecasting with {3D} neural
  networks.
\newblock \emph{Nature}, 619\penalty0 (7970):\penalty0 533--538, 2023.

\bibitem[Bodnar et~al.(2025)Bodnar, Bruinsma, Lucic, Stanley, Allen,
  Brandstetter, Garvan, Riechert, Weyn, Dong, Gupta, Thambiratnam, Archibald,
  Wu, Heider, Welling, Turner, and Perdikaris]{bodnarFoundationModelEarth2025a}
Cristian Bodnar, Wessel~P. Bruinsma, Ana Lucic, Megan Stanley, Anna Allen,
  Johannes Brandstetter, Patrick Garvan, Maik Riechert, Jonathan~A. Weyn, Haiyu
  Dong, Jayesh~K. Gupta, Kit Thambiratnam, Alexander~T. Archibald, Chun-Chieh
  Wu, Elizabeth Heider, Max Welling, Richard~E. Turner, and Paris Perdikaris.
\newblock A foundation model for the {Earth} system.
\newblock \emph{Nature}, 641\penalty0 (8065):\penalty0 1180--1187, 2025.

\bibitem[Bonev et~al.(2023)Bonev, Kurth, Hundt, Pathak, Baust, Kashinath, and
  Anandkumar]{bonevSphericalFourierNeural2023}
Boris Bonev, Thorsten Kurth, Christian Hundt, Jaideep Pathak, Maximilian Baust,
  Karthik Kashinath, and Anima Anandkumar.
\newblock {Spherical Fourier Neural Operators}: Learning stable dynamics on the
  sphere.
\newblock In \emph{Proceedings of the 40th International Conference on Machine
  Learning}, pages 2806--2823, 2023.

\bibitem[Bonev et~al.(2025{\natexlab{a}})Bonev, Kurth, Mahesh, Bisson,
  Kossaifi, Kashinath, Anandkumar, Collins, Pritchard, and
  Keller]{bonevFourCastNet3Geometric2025}
Boris Bonev, Thorsten Kurth, Ankur Mahesh, Mauro Bisson, Jean Kossaifi, Karthik
  Kashinath, Anima Anandkumar, William~D. Collins, Michael~S. Pritchard, and
  Alexander Keller.
\newblock {FourCastNet} 3: A geometric approach to probabilistic
  machine-learning weather forecasting at scale.
\newblock arXiv preprint arXiv:2507.12144, 2025{\natexlab{a}}.

\bibitem[Bonev et~al.(2025{\natexlab{b}})Bonev, Rietmann, Paris, Carpentieri,
  and Kurth]{bonevAttentionSphere2025}
Boris Bonev, Max Rietmann, Andrea Paris, Alberto Carpentieri, and Thorsten
  Kurth.
\newblock Attention on the sphere.
\newblock In \emph{Advances in Neural Information Processing Systems}, pages
  61662--61688, 2025{\natexlab{b}}.

\bibitem[Camps-Valls et~al.(2025)Camps-Valls, Fern{\'{a}}ndez-Torres, Cohrs,
  H{\"{o}}hl, Castelletti, Pacal, Robin, Martinuzzi, Papoutsis, Prapas,
  P{\'{e}}rez-Aracil, Weigel, Gonzalez-Calabuig, Reichstein, Rabel, Giuliani,
  Mahecha, Popescu, Pellicer-Valero, Ouala, Salcedo-Sanz, Sippel, Kondylatos,
  Happ{\'{e}}, and Williams]{camps-vallsArtificialIntelligenceModeling2025}
Gustau Camps-Valls, Miguel-{\'{A}}ngel Fern{\'{a}}ndez-Torres, Kai-Hendrik
  Cohrs, Adrian H{\"{o}}hl, Andrea Castelletti, Aytac Pacal, Claire Robin,
  Francesco Martinuzzi, Ioannis Papoutsis, Ioannis Prapas, Jorge
  P{\'{e}}rez-Aracil, Katja Weigel, Maria Gonzalez-Calabuig, Markus Reichstein,
  Martin Rabel, Matteo Giuliani, Miguel~D. Mahecha, Oana-Iuliana Popescu,
  Oscar~J. Pellicer-Valero, Said Ouala, Sancho Salcedo-Sanz, Sebastian Sippel,
  Spyros Kondylatos, Tamara Happ{\'{e}}, and Tristan Williams.
\newblock Artificial intelligence for modeling and understanding extreme
  weather and climate events.
\newblock \emph{Nature Communications}, 16\penalty0 (1):\penalty0 1919, 2025.

\bibitem[Chen et~al.(2025)Chen, Han, Ling, Gong, Bai, Wang, Luo, Fei, Zhang,
  Chen, Ma, Zhang, Su, Ci, Li, Yang, and Ouyang]{chenFengWuPushingSkillful2023}
Kang Chen, Tao Han, Fenghua Ling, Junchao Gong, Lei Bai, Xinyu Wang, Jing-Jia
  Luo, Ben Fei, Wenlong Zhang, Xi Chen, Leiming Ma, Tianning Zhang, Rui Su,
  Yuanzheng Ci, Bin Li, Xiaokang Yang, and Wanli Ouyang.
\newblock The operational medium-range deterministic weather forecasting can be
  extended beyond a 10-day lead time.
\newblock \emph{Communications Earth \& Environment}, 6\penalty0 (1):\penalty0
  518, 2025.

\bibitem[Chen et~al.(2023)Chen, Zhong, Zhang, Cheng, Xu, Qi, and
  Li]{chenFuXiCascadeMachine2023}
Lei Chen, Xiaohui Zhong, Feng Zhang, Yuan Cheng, Yinghui Xu, Yuan Qi, and Hao
  Li.
\newblock {FuXi}: A cascade machine learning forecasting system for 15-day
  global weather forecast.
\newblock \emph{npj Climate and Atmospheric Science}, 6\penalty0 (1):\penalty0
  190, 2023.

\bibitem[Chen et~al.(2024)Chen, Zhong, Li, Wu, Lu, Chen, Xie, Wu, Chao, Lin,
  Hu, and Qi]{chenMachineLearningModel2024}
Lei Chen, Xiaohui Zhong, Hao Li, Jie Wu, Bo Lu, Deliang Chen, Shang-Ping Xie,
  Libo Wu, Qingchen Chao, Chensen Lin, Zixin Hu, and Yuan Qi.
\newblock A machine learning model that outperforms conventional global
  subseasonal forecast models.
\newblock \emph{Nature Communications}, 15\penalty0 (1):\penalty0 6425, 2024.

\bibitem[Couairon et~al.(2026)Couairon, Singh, Charantonis, Lessig, and
  Monteleoni]{couaironArchesWeatherGenSkillfulComputeefficient2026}
Guillaume Couairon, Renu Singh, Anastase Charantonis, Christian Lessig, and
  Claire Monteleoni.
\newblock {ArchesWeatherGen}: Skillful and compute-efficient probabilistic
  weather forecasting with machine learning.
\newblock \emph{Science Advances}, 12\penalty0 (17):\penalty0 eadx2372, 2026.

\bibitem[Derber and Bouttier(1999)]{derberReformulationBackgroundError1999}
J. Derber and F. Bouttier.
\newblock A reformulation of the background error covariance in the {ECMWF}
  global data assimilation system.
\newblock \emph{Tellus A: Dynamic Meteorology and Oceanography}, 51\penalty0
  (2):\penalty0 195--221, 1999.

\bibitem[Geng et~al.(2025)Geng, Deng, Bai, Kolter, and
  He]{gengMeanFlowsOnestep}
Zhengyang Geng, Mingyang Deng, Xingjian Bai, Zico Kolter, and Kaiming He.
\newblock {Mean Flows} for one-step generative modeling.
\newblock In \emph{Advances in Neural Information Processing Systems}, pages
  75460--75482, 2025.

\bibitem[Geng et~al.(2026)Geng, Lu, Wu, Shechtman, Kolter, and
  He]{gengImprovedMeanFlows}
Zhengyang Geng, Yiyang Lu, Zongze Wu, Eli Shechtman, J.~Zico Kolter, and
  Kaiming He.
\newblock Improved {Mean Flows}: On the challenges of fastforward generative
  models.
\newblock In \emph{Proceedings of the IEEE/CVF Conference on Computer Vision
  and Pattern Recognition (CVPR)}, pages 30467--30476, 2026.

\bibitem[Gneiting and Raftery(2007)]{gneitingStrictlyProperScoring2007}
Tilmann Gneiting and Adrian~E. Raftery.
\newblock Strictly proper scoring rules, prediction, and estimation.
\newblock \emph{Journal of the American Statistical Association}, 102\penalty0
  (477):\penalty0 359--378, 2007.

\bibitem[Guo et~al.(2024)Guo, Zhou, Jiang, Wu, Sun, and
  Jin]{guoMaximizingImpactDeep2024}
Yizhen Guo, Tian Zhou, Wanyi Jiang, Bo Wu, Liang Sun, and Rong Jin.
\newblock Maximizing the impact of deep learning on subseasonal-to-seasonal
  climate forecasting: The essential role of optimization.
\newblock arXiv preprint arXiv:2411.16728, 2024.

\bibitem[Hersbach et~al.(2020)Hersbach, Bell, Berrisford, Hirahara,
  Hor{\'{a}}nyi, Mu{\~{n}}oz-Sabater, Nicolas, Peubey, Radu, Schepers, Simmons,
  Soci, Abdalla, Abellan, Balsamo, Bechtold, Biavati, Bidlot, Bonavita,
  De~Chiara, Dahlgren, Dee, Diamantakis, Dragani, Flemming, Forbes, Fuentes,
  Geer, Haimberger, Healy, Hogan, H{\'{o}}lm, Janiskov{\'{a}}, Keeley,
  Laloyaux, Lopez, Lupu, Radnoti, de~Rosnay, Rozum, Vamborg, Villaume, and
  Th{\'{e}}paut]{hersbachERA5GlobalReanalysis2020}
Hans Hersbach, Bill Bell, Paul Berrisford, Shoji Hirahara, Andr{\'{a}}s
  Hor{\'{a}}nyi, Joaqu{\'{i}}n Mu{\~{n}}oz-Sabater, Julien Nicolas, Carole
  Peubey, Raluca Radu, Dinand Schepers, Adrian Simmons, Cornel Soci, Saleh
  Abdalla, Xavier Abellan, Gianpaolo Balsamo, Peter Bechtold, Gionata Biavati,
  Jean Bidlot, Massimo Bonavita, Giovanna De~Chiara, Per Dahlgren, Dick Dee,
  Michail Diamantakis, Rossana Dragani, Johannes Flemming, Richard Forbes,
  Manuel Fuentes, Alan Geer, Leo Haimberger, Sean Healy, Robin~J. Hogan,
  El{\'{i}}as H{\'{o}}lm, Marta Janiskov{\'{a}}, Sarah Keeley, Patrick
  Laloyaux, Philippe Lopez, Cristina Lupu, Gabor Radnoti, Patricia de Rosnay,
  Iryna Rozum, Freja Vamborg, Sebastien Villaume, and Jean-No{\"{e}}l
  Th{\'{e}}paut.
\newblock The {ERA5} global reanalysis.
\newblock \emph{Quarterly Journal of the Royal Meteorological Society},
  146\penalty0 (730):\penalty0 1999--2049, 2020.

\bibitem[Kettleborough et~al.(2026)Kettleborough, Davis, Comer, Martin,
  Barbosa~Aguiar, Collier, Scaife, Clark, Mancell, Smout-Day, and
  Hardiman]{kettleboroughGlobalSeasonalForecasting2025}
J.~A. Kettleborough, P.~J. Davis, R.~E. Comer, M.~J. Martin, A. Barbosa~Aguiar,
  T. Collier, A.~A. Scaife, A. Clark, J. Mancell, K.~A. Smout-Day, and S.~C.
  Hardiman.
\newblock {Global Seasonal Forecasting System 6 (GloSea6)}: A large ensemble
  seasonal forecasting system.
\newblock \emph{Monthly Weather Review}, 154\penalty0 (1):\penalty0 39--57,
  2026.

\bibitem[Kovachki et~al.(2023)Kovachki, Li, Liu, Azizzadenesheli, Bhattacharya,
  Stuart, and Anandkumar]{kovachkiNeuralOperatorLearning2023}
Nikola Kovachki, Zongyi Li, Burigede Liu, Kamyar Azizzadenesheli, Kaushik
  Bhattacharya, Andrew Stuart, and Anima Anandkumar.
\newblock {Neural Operator}: Learning maps between function spaces with
  applications to {PDEs}.
\newblock \emph{Journal of Machine Learning Research}, 24\penalty0
  (89):\penalty0 1--97, 2023.

\bibitem[Lam et~al.(2023)Lam, Sanchez-Gonzalez, Willson, Wirnsberger,
  Fortunato, Alet, Ravuri, Ewalds, Eaton-Rosen, Hu, Merose, Hoyer, Holland,
  Vinyals, Stott, Pritzel, Mohamed, and
  Battaglia]{lamLearningSkillfulMediumrange2023}
Remi Lam, Alvaro Sanchez-Gonzalez, Matthew Willson, Peter Wirnsberger, Meire
  Fortunato, Ferran Alet, Suman Ravuri, Timo Ewalds, Zach Eaton-Rosen, Weihua
  Hu, Alexander Merose, Stephan Hoyer, George Holland, Oriol Vinyals, Jacklynn
  Stott, Alexander Pritzel, Shakir Mohamed, and Peter Battaglia.
\newblock Learning skillful medium-range global weather forecasting.
\newblock \emph{Science}, 382\penalty0 (6677):\penalty0 1416--1421, 2023.

\bibitem[Lang et~al.(2026)Lang, Alexe, Clare, Roberts, Adewoyin,
  Ben~Bouall{\`{e}}gue, Chantry, Dramsch, Dueben, Hahner, Maciel,
  Prieto-Nemesio, O'Brien, Pinault, Polster, Raoult, Tietsche, and
  Leutbecher]{langAIFSCRPSEnsembleForecasting2026}
Simon Lang, Mihai Alexe, Mariana C.~A. Clare, Christopher Roberts, Rilwan
  Adewoyin, Zied Ben~Bouall{\`{e}}gue, Matthew Chantry, Jesper Dramsch,
  Peter~D. Dueben, Sara Hahner, Pedro Maciel, Ana Prieto-Nemesio, Cathal
  O'Brien, Florian Pinault, Jan Polster, Baudouin Raoult, Steffen Tietsche, and
  Martin Leutbecher.
\newblock {AIFS-CRPS}: Ensemble forecasting using a model trained with a loss
  function based on the continuous ranked probability score.
\newblock \emph{npj Artificial Intelligence}, 2\penalty0 (1):\penalty0 18,
  2026.

\bibitem[Li et~al.(2026)Li, Liu, Liu, Chen, Cao, Wang, Zheng, Zhang, Liang,
  Zhang, Wang, Jin, Cheng, and
  Fu]{liTIANQUANS2SSUBSEASONALTOSEASONALGLOBAL2026}
Guowen Li, Xintong Liu, Yang Liu, Mengxuan Chen, Shilei Cao, Xuehe Wang,
  Juepeng Zheng, Jinxiao Zhang, Haoyuan Liang, Lixian Zhang, Jiuke Wang, Meng
  Jin, Hong Cheng, and Haohuan Fu.
\newblock {TianQuan-S2S}: A subseasonal-to-seasonal global weather model via
  incorporate climatology state.
\newblock In \emph{International Conference on Learning Representations}, pages
  106063--106088, 2026.

\bibitem[Li et~al.(2021)Li, Kovachki, Azizzadenesheli, Liu, Bhattacharya,
  Stuart, and Anandkumar]{liFourierNeuralOperator2021}
Zongyi Li, Nikola Kovachki, Kamyar Azizzadenesheli, Burigede Liu, Kaushik
  Bhattacharya, Andrew Stuart, and Anima Anandkumar.
\newblock {Fourier Neural Operator} for parametric partial differential
  equations.
\newblock In \emph{International Conference on Learning Representations}, 2021.

\bibitem[Lipman et~al.(2023)Lipman, Chen, Ben-Hamu, Nickel, and
  Le]{lipmanFlowMatchingGenerative2023}
Yaron Lipman, Ricky T.~Q. Chen, Heli Ben-Hamu, Maximilian Nickel, and Matt Le.
\newblock Flow matching for generative modeling.
\newblock In \emph{International Conference on Learning Representations}, 2023.

\bibitem[Liu et~al.(2025)Liu, Zheng, Cheng, Tsung, Zhao, Rong, and
  Li]{liuCirTGlobalSubseasonaltoSeasonal2025a}
Yang Liu, Zinan Zheng, Jiashun Cheng, Fugee Tsung, Deli Zhao, Yu Rong, and Jia
  Li.
\newblock {CirT}: Global subseasonal-to-seasonal forecasting with
  geometry-inspired transformer.
\newblock In \emph{International Conference on Learning Representations}, pages
  101845--101866, 2025.

\bibitem[Liu-Schiaffini et~al.(2024)Liu-Schiaffini, Berner, Bonev, Kurth,
  Azizzadenesheli, and Anandkumar]{liu-schiaffiniNeuralOperatorsLocalized2024}
Miguel Liu-Schiaffini, Julius Berner, Boris Bonev, Thorsten Kurth, Kamyar
  Azizzadenesheli, and Anima Anandkumar.
\newblock Neural operators with localized integral and differential kernels.
\newblock In \emph{Proceedings of the 41st International Conference on Machine
  Learning}, pages 32576--32594, 2024.

\bibitem[Loshchilov and Hutter(2019)]{loshchilovDecoupledWeightDecay2019}
Ilya Loshchilov and Frank Hutter.
\newblock Decoupled weight decay regularization.
\newblock In \emph{International Conference on Learning Representations}, 2019.

\bibitem[Lyu et~al.(2026)Lyu, Zhang, and
  Liu]{lyuPhysicsInformedTeleconnectionAwareTransformer2025}
Tengfei Lyu, Weijia Zhang, and Hao Liu.
\newblock Physics-informed teleconnection-aware transformer for global
  subseasonal-to-seasonal forecasting.
\newblock In \emph{Proceedings of the 32nd ACM SIGKDD Conference on Knowledge
  Discovery and Data Mining V.1}, pages 1054--1065, 2026.

\bibitem[Merryfield et~al.(2020)Merryfield, Baehr, Batt{\'{e}}, Becker, Butler,
  Coelho, Danabasoglu, Dirmeyer, Doblas-Reyes, Domeisen, Ferranti, Ilynia,
  Kumar, M{\"{u}}ller, Rixen, Robertson, Smith, Takaya, Tuma, Vitart, White,
  Alvarez, Ardilouze, Attard, Baggett, Balmaseda, Beraki, Bhattacharjee,
  Bilbao, de~Andrade, DeFlorio, D{\'{i}}az, Ehsan, Fragkoulidis, Gonzalez,
  Grainger, Green, Hell, Infanti, Isensee, Kataoka, Kirtman, Klingaman, Lee,
  Mayer, McKay, Mecking, Miller, Neddermann, Ng, Oss{\'{o}}, Pankatz, Peatman,
  Pegion, Perlwitz, Recalde-Coronel, Reintges, Renkl, Solaraju-Murali, Spring,
  Stan, Sun, Tozer, Vigaud, Woolnough, and
  Yeager]{merryfieldCurrentEmergingDevelopments2020}
William~J. Merryfield, Johanna Baehr, Lauriane Batt{\'{e}}, Emily~J. Becker,
  Amy~H. Butler, Caio A.~S. Coelho, Gokhan Danabasoglu, Paul~A. Dirmeyer,
  Francisco~J. Doblas-Reyes, Daniela I.~V. Domeisen, Laura Ferranti, Tatiana
  Ilynia, Arun Kumar, Wolfgang~A. M{\"{u}}ller, Michel Rixen, Andrew~W.
  Robertson, Doug~M. Smith, Yuhei Takaya, Matthias Tuma, Frederic Vitart,
  Christopher~J. White, Mariano~S. Alvarez, Constantin Ardilouze, Hannah
  Attard, Cory Baggett, Magdalena~A. Balmaseda, Asmerom~F. Beraki, Partha~S.
  Bhattacharjee, Roberto Bilbao, Felipe~M. de Andrade, Michael~J. DeFlorio,
  Leandro~B. D{\'{i}}az, Muhammad~Azhar Ehsan, Georgios Fragkoulidis, Alex~O.
  Gonzalez, Sam Grainger, Benjamin~W. Green, Momme~C. Hell, Johnna~M. Infanti,
  Katharina Isensee, Takahito Kataoka, Ben~P. Kirtman, Nicholas~P. Klingaman,
  June-Yi Lee, Kirsten Mayer, Roseanna McKay, Jennifer~V. Mecking, Douglas~E.
  Miller, Nele Neddermann, Ching Ho~Justin Ng, Albert Oss{\'{o}}, Klaus
  Pankatz, Simon Peatman, Kathy Pegion, Judith Perlwitz, G.~Cristina
  Recalde-Coronel, Annika Reintges, Christoph Renkl, Balakrishnan
  Solaraju-Murali, Aaron Spring, Cristiana Stan, Y.~Qiang Sun, Carly~R. Tozer,
  Nicolas Vigaud, Steven Woolnough, and Stephen Yeager.
\newblock Current and emerging developments in subseasonal to decadal
  prediction.
\newblock \emph{Bulletin of the American Meteorological Society}, 101\penalty0
  (6):\penalty0 E869--E896, 2020.

\bibitem[Molteni et~al.(1996)Molteni, Buizza, Palmer, and
  Petroliagis]{molteniECMWFEnsemblePrediction1996}
F. Molteni, R. Buizza, T.~N. Palmer, and T. Petroliagis.
\newblock The {ECMWF} ensemble prediction system: Methodology and validation.
\newblock \emph{Quarterly Journal of the Royal Meteorological Society},
  122\penalty0 (529):\penalty0 73--119, 1996.

\bibitem[Nathaniel et~al.(2024)Nathaniel, Qu, Nguyen, Yu, Busecke, Grover, and
  Gentine]{nathanielChaosBenchMultiChannelPhysicsBased2024}
Juan Nathaniel, Yongquan Qu, Tung Nguyen, Sungduk Yu, Julius Busecke, Aditya
  Grover, and Pierre Gentine.
\newblock {ChaosBench}: A multi-channel, physics-based benchmark for
  subseasonal-to-seasonal climate prediction.
\newblock In \emph{Advances in Neural Information Processing Systems}, pages
  43715--43729, 2024.

\bibitem[Nguyen et~al.(2025)Nguyen, Pham, Arcomano, Kotamarthi, Foster,
  Madireddy, and Grover]{nguyenOmniCastMaskedLatent2025}
Tung Nguyen, Tuan Pham, Troy Arcomano, Rao Kotamarthi, Ian Foster, Sandeep
  Madireddy, and Aditya Grover.
\newblock {OmniCast}: A masked latent diffusion model for weather forecasting
  across time scales.
\newblock In \emph{Advances in Neural Information Processing Systems}, pages
  172182--172212, 2025.

\bibitem[Pathak et~al.(2022)Pathak, Subramanian, Harrington, Raja,
  Chattopadhyay, Mardani, Kurth, Hall, Li, Azizzadenesheli, Hassanzadeh,
  Kashinath, and Anandkumar]{pathakFourCastNetGlobalDatadriven2022}
Jaideep Pathak, Shashank Subramanian, Peter Harrington, Sanjeev Raja, Ashesh
  Chattopadhyay, Morteza Mardani, Thorsten Kurth, David Hall, Zongyi Li, Kamyar
  Azizzadenesheli, Pedram Hassanzadeh, Karthik Kashinath, and Animashree
  Anandkumar.
\newblock {FourCastNet}: A global data-driven high-resolution weather model
  using adaptive {Fourier} neural operators.
\newblock arXiv preprint arXiv:2202.11214, 2022.

\bibitem[Perez et~al.(2018)Perez, Strub, De~Vries, Dumoulin, and
  Courville]{perezFiLMVisualReasoning2018}
Ethan Perez, Florian Strub, Harm De~Vries, Vincent Dumoulin, and Aaron
  Courville.
\newblock {FiLM}: Visual reasoning with a general conditioning layer.
\newblock In \emph{Proceedings of the AAAI Conference on Artificial
  Intelligence}, 2018.

\bibitem[Price et~al.(2025)Price, Sanchez-Gonzalez, Alet, Andersson, El-Kadi,
  Masters, Ewalds, Stott, Mohamed, Battaglia, Lam, and
  Willson]{priceProbabilisticWeatherForecasting2025}
Ilan Price, Alvaro Sanchez-Gonzalez, Ferran Alet, Tom~R. Andersson, Andrew
  El-Kadi, Dominic Masters, Timo Ewalds, Jacklynn Stott, Shakir Mohamed, Peter
  Battaglia, Remi Lam, and Matthew Willson.
\newblock Probabilistic weather forecasting with machine learning.
\newblock \emph{Nature}, 637\penalty0 (8044):\penalty0 84--90, 2025.

\bibitem[Rasp et~al.(2024)Rasp, Hoyer, Merose, Langmore, Battaglia, Russell,
  Sanchez-Gonzalez, Yang, Carver, Agrawal, Chantry, Ben~Bouall{\`{e}}gue,
  Dueben, Bromberg, Sisk, Barrington, Bell, and
  Sha]{raspWeatherBench2Benchmark2024a}
Stephan Rasp, Stephan Hoyer, Alexander Merose, Ian Langmore, Peter Battaglia,
  Tyler Russell, Alvaro Sanchez-Gonzalez, Vivian Yang, Rob Carver, Shreya
  Agrawal, Matthew Chantry, Zied Ben~Bouall{\`{e}}gue, Peter Dueben, Carla
  Bromberg, Jared Sisk, Luke Barrington, Aaron Bell, and Fei Sha.
\newblock {WeatherBench} 2: A benchmark for the next generation of data-driven
  global weather models.
\newblock \emph{Journal of Advances in Modeling Earth Systems}, 16\penalty0
  (6):\penalty0 e2023MS004019, 2024.

\bibitem[Rasp et~al.(2026)Rasp, Babenko, Masters, El-Kadi, Merchant, Shalev,
  Price, Zyda, Lam, Shysheya, Willson, Markou, Agrawal, Vora, Hassen, Mak,
  Andersson, Bela, Uddin, Levi, Gaiarin, Alet, Bell, Battaglia, and
  Sanchez-Gonzalez]{raspWeatherNext3Increasing2026a}
Stephan Rasp, Boris Babenko, Dominic Masters, Andrew El-Kadi, Samier Merchant,
  Guy Shalev, Ilan Price, Fred Zyda, Remi Lam, Sasha Shysheya, Matthew Willson,
  Stratis Markou, Shreya Agrawal, Suhani Vora, Mohammed~Alewi Hassen, Sunny
  Mak, Tom~R. Andersson, Megan Bela, Akib Uddin, Nofar~Peled Levi, Ben Gaiarin,
  Ferran Alet, Aaron Bell, Peter Battaglia, and Alvaro Sanchez-Gonzalez.
\newblock {WeatherNext} 3: Increasing resolution and performance of global
  weather models with raw observations.
\newblock arXiv preprint arXiv:2609.03582, 2026.

\bibitem[R{\"u}hling~Cachay et~al.(2026)R{\"u}hling~Cachay, Watson-Parris, and
  Yu]{cachayUCastSurprisinglySimple2026}
Salva R{\"u}hling~Cachay, Duncan Watson-Parris, and Rose Yu.
\newblock {U-Cast}: A surprisingly simple and efficient frontier probabilistic
  {AI} weather forecaster.
\newblock In \emph{Proceedings of the 43rd International Conference on Machine
  Learning}, pages 106213--106234, 2026.

\bibitem[Saha et~al.(2014)Saha, Moorthi, Wu, Wang, Nadiga, Tripp, Behringer,
  Hou, Chuang, Iredell, Ek, Meng, Yang, Mendez, van~den Dool, Zhang, Wang,
  Chen, and Becker]{sahaNCEPClimateForecast2014}
Suranjana Saha, Shrinivas Moorthi, Xingren Wu, Jiande Wang, Sudhir Nadiga,
  Patrick Tripp, David Behringer, Yu-Tai Hou, Hui-ya Chuang, Mark Iredell,
  Michael Ek, Jesse Meng, Rongqian Yang, Malaqu{\'{i}}as~Pe{\~{n}}a Mendez,
  Huug van~den Dool, Qin Zhang, Wanqiu Wang, Mingyue Chen, and Emily Becker.
\newblock The {NCEP Climate Forecast System Version 2}.
\newblock \emph{Journal of Climate}, 27\penalty0 (6):\penalty0 2185--2208,
  2014.

\bibitem[Vitart(2014)]{vitartEvolutionECMWFSubseasonal2014}
Fr{\'{e}}d{\'{e}}ric Vitart.
\newblock Evolution of {ECMWF} sub-seasonal forecast skill scores.
\newblock \emph{Quarterly Journal of the Royal Meteorological Society},
  140\penalty0 (683):\penalty0 1889--1899, 2014.

\bibitem[Xu et~al.(2026)Xu, Gao, Wang, Su, Ling, Wu, and
  Ouyang]{xuTycheOneStep2026}
Fan Xu, Yuan Gao, Kun Wang, Rui Su, Fenghua Ling, Hao Wu, and Wanli Ouyang.
\newblock Tyche: One step flow for efficient probabilistic weather forecasting.
\newblock In \emph{Advances in Neural Information Processing Systems}, 2026.

\end{thebibliography}
}

\makeatletter\renewcommand*{\Hy@backout}[1]{}\makeatother 

\clearpage
\maketitlesupplementary
\appendix


\section{Model Details}
\label{sec:supp_model}

\subsection{Network architecture}
\label{sec:supp_architecture}

\paragraph{Spherical harmonic transform.}
On an $H \times W$ spherical grid, the SHT projects a field $f$ onto the spherical harmonic basis $Y_\ell^m(\vartheta, \lambda) = \bar{P}_\ell^m(\cos\vartheta)\, e^{\mathrm{i} m \lambda}$ as
\begin{equation}
	\hat{f}_\ell^m = \sum_{i=1}^{H} q_i\, \bar{P}_\ell^m(\cos\vartheta_i)\, \frac{2\pi}{W} \sum_{j=1}^{W} f(\vartheta_i, \lambda_j)\, e^{-\mathrm{i} m \lambda_j},
	\label{eq:supp_sht}
\end{equation}
where $\bar{P}_\ell^m$ is the normalized associated Legendre function and $q_i$ are the quadrature weights at colatitudes $\vartheta_i$.
The inverse SHT reconstructs the spatial field by evaluating the spherical harmonic expansion on the grid.
In our model, both networks place their latent grid latitudes at Gauss--Legendre nodes.

\paragraph{Spectral convolution.}
By the spherical convolution theorem, convolving a spherical field with an axisymmetric kernel scales each coefficient $\hat{f}_\ell^m$ by a factor that depends only on its degree $\ell$.
SFNO~\citep{bonevSphericalFourierNeural2023} generalizes these scalar factors to learnable complex matrices $W_\ell \in \mathbb{C}^{D \times D}$ that mix the $D$ latent channels,
\begin{equation}
	\mathcal{K}_{\mathrm{G}}\, h = \mathrm{SHT}^{-1}\big[\, W_\ell\, \mathrm{SHT}(h)_\ell^m \,\big].
	\label{eq:supp_sfno}
\end{equation}
Sharing $W_\ell$ across zonal orders $m$ makes $\mathcal{K}_{\mathrm{G}}$ equivariant to sphere rotations and gives every output point a global receptive field.

\paragraph{DISCO convolution.}
Discrete--continuous (DISCO) convolution~\citep{liu-schiaffiniNeuralOperatorsLocalized2024} pairs a locally supported kernel with spherical quadrature.
A continuous filter on a geodesic disk of radius $\theta_c$ is expanded in a fixed basis $\{\psi_n\}$, and the convolution integral is discretized over input grid points $x_j$ as
\begin{equation}
	(\mathcal{K}_{\mathrm{L}}\, h)(x) = \sum_{n} W_n \sum_{j} \psi_n\big(R_{x}^{-1} x_j\big)\, h(x_j)\, q_j,
	\label{eq:supp_disco}
\end{equation}
where $R_{x}$ rotates the north pole to the output location $x$, $q_j$ are quadrature weights, and $W_n \in \mathbb{R}^{D' \times D}$ are learnable weights.
Evaluating the basis at true geodesic offsets keeps the physical filter footprint identical across latitudes despite meridian convergence near the poles.
Because the tensor $\psi_n(R_{x}^{-1} x_j)\, q_j$ is sparse and precomputed, the operation reduces to sparse matrix multiplication and naturally handles grid resampling between input and output grids.
We adopt a nine-function Morlet basis formed by multiplying tangent-plane Fourier modes along two orthogonal axes with a Hann window that vanishes at $\theta_c$.

\begin{table}[t]
	\caption{
		Architecture of the deterministic backbone $F_\theta$ and the one-step generator $G_\phi$.
		G and L denote global SFNO and local DISCO blocks, and DOY denotes the day of year.
	}
	\label{tab:supp_architecture}
	\centering
	\small
	\begin{tabular*}{\columnwidth}{@{\extracolsep{\fill}}lcc@{}}
		\toprule
		\textbf{Setting} & \boldmath$F_\theta$ & \boldmath$G_\phi$ \\
		\midrule
		Input / output channels & 59 / 59 & 177 / 59 \\
		Latent grid & $60 \times 120$ & $60 \times 120$ \\
		Latent channels & $30 \times 10 + 36$ & $25 \times 10 + 45$ \\
		Processor blocks & $3 \times$ GLL & $2 \times$ GLL \\
		MLP ratio / activation & 2 / GELU & 2 / GELU \\
		Normalization & Instance norm & Instance norm \\
		FiLM conditioning & DOY, lead & DOY, lead, $s'$, $s$ \\
		Global skip connection & Yes & No \\
		Parameters & 30.6\,M & 16.4\,M \\
		\bottomrule
	\end{tabular*}
\end{table}

\paragraph{Encoder, processor, and decoder.}
Table~\ref{tab:supp_architecture} summarizes the architectural settings of both networks.
The encoder downsamples the $121 \times 240$ equiangular input grid to the $60 \times 120$ latent grid using dedicated level-wise modules.
Each of the ten pressure levels has its own encoder, applying a DISCO convolution with $\theta_c = 3^\circ$ and a pointwise MLP to its five variables, and an identical module embeds the nine surface variables together with the land--sea mask.
In $G_\phi$, each level encoder receives the matching variables from $z_s$, $X_t$, and $\tilde{X}_{t+1}$ jointly, embedding the noisy residual alongside the current and predicted states.
Concatenating all level embeddings produces 336 latent channels for $F_\theta$ and 295 for $G_\phi$.
Each processor block updates the latent field $h$ via
\begin{equation}
	h \leftarrow h + \gamma \odot \mathrm{MLP}\Big(\mathrm{FiLM}\big(\mathrm{Norm}(\mathcal{K} h)\big)\Big),
	\label{eq:supp_block}
\end{equation}
where $\mathcal{K}$ is $\mathcal{K}_{\mathrm{G}}$ in a global block and $\mathcal{K}_{\mathrm{L}}$ in a local block, $\mathrm{Norm}$ denotes instance normalization over the latent grid, $\mathrm{MLP}$ is a pointwise two-layer GELU perceptron with expansion ratio 2, and $\gamma$ is a per-channel layer scale initialized to 0.1.
The processor stacks repeating groups of one global block and two local blocks, three times in $F_\theta$ and twice in $G_\phi$.
Mirroring the encoder, each level-specific decoder applies a pointwise MLP, bilinear upsampling to the $121 \times 240$ grid, and a DISCO convolution with $\theta_c = 3^\circ$.
$F_\theta$ adds a learned $1 {\times} 1$ projection of its input state to the decoder output as a global skip connection.
$G_\phi$ omits this skip connection because it predicts a velocity field for the residual flow.

\paragraph{Conditioning.}
Every processor block injects a conditioning vector $c$ via feature-wise linear modulation (FiLM)~\citep{perezFiLMVisualReasoning2018}, computing $(1 + a) \odot h + b$ from a zero-initialized linear projection of $c$ so each block starts unconditioned.
Both networks encode the valid day of year cyclically using sine and cosine features and normalize the lead day $k \in \{1, \dots, 44\}$ to $[0, 1]$.
$F_\theta$ feeds these calendar and lead features directly into FiLM, allowing the backbone to adapt as autoregressive states drift from reanalysis statistics.
$G_\phi$ sums 128-dimensional MLP embeddings of the flow times $s$ and $s'$ (via sinusoidal embeddings $e_s$ and $e_{s'}$), the day of year, and the lead day ($e_\kappa$).

\paragraph{Positional embedding.}
Both spectral and DISCO convolutions act uniformly across the sphere.
To capture location-dependent relationships, we supply the land--sea mask and add a learned positional embedding to the latent features after the encoder.
For latitude $\varphi = \pi/2 - \vartheta$ and longitude $\lambda$ in radians, the positional vector $\mathbf{p}$ concatenates normalized latitude with Fourier features $\mathbf{f}_j$ at harmonics $j = 1, \ldots, 4$,
\begin{equation*}
\begin{aligned}
	\mathbf{p} &= \big[2\varphi/\pi,\, \mathbf{f}_1,\, \ldots,\, \mathbf{f}_4\big], \\
	\mathbf{f}_j &= \big[\sin(j\varphi),\, \cos(j\varphi),\, \sin(j\lambda),\, \cos(j\lambda)\big].
\end{aligned}
\end{equation*}
A zero-initialized $1 \times 1$ convolution projects these 17 features to the latent channels at each grid point.

\subsection{One-step generation}
\label{sec:supp_imf}

\paragraph{Interval sampling and loss weight.}
During training, we draw two times from a logit-normal distribution with location $-0.4$ and scale $1.0$, assign the smaller to $s'$ and the larger to $s$, and set $s' = s$ for half of each mini-batch.
When $s' = s$, the factor $s - s'$ eliminates the derivative term in Eq.~\eqref{eq:imf}, reducing the loss to standard flow matching on the instantaneous velocity $v_\phi(z_s, s) = u_\phi(z_s, s, s)$.
Following MeanFlow~\citep{gengMeanFlowsOnestep}, we compute the per-sample adaptive weight as
\begin{equation}
	w = \mathrm{sg}\Big[\big(\big\| V_\phi(z_s, s', s) - (\varepsilon - r) \big\|_{A}^{2} + 10^{-3}\big)^{-1/2}\Big].
	\label{eq:imf_weight}
\end{equation}
This weighting makes the objective scale like an $L_2$ norm and keeps occasional derivative spikes from dominating the gradients.

\paragraph{Computing the total derivative.}
By the chain rule, the total derivative in Eq.~\eqref{eq:imf} expands to
\begin{equation}
	\frac{\mathrm{d}}{\mathrm{d}s} u_\phi(z_s, s', s) = \frac{\partial u_\phi}{\partial z_s}\, v_\phi + \frac{\partial u_\phi}{\partial s},
	\label{eq:supp_jvp}
\end{equation}
where the instantaneous velocity $v_\phi$ acts as the rate of change of $z_s$ and is treated as a constant via stop-gradient.
Computing the full Jacobian $\partial u_\phi / \partial z_s$ explicitly is intractable for high-dimensional weather fields, so we evaluate the total derivative as a Jacobian--vector product (JVP) using forward-mode automatic differentiation.
Intuitively, forward-mode differentiation runs alongside the regular forward pass and tracks both each layer's activation $h$ and its directional derivative $\dot{h}$ with respect to $s$ in one sweep.
Samples with $s' = s$ need only a standard forward pass because the derivative term vanishes.
For samples with $s' < s$, we first compute $v_\phi$ without tracking gradients, and then run a single forward-mode pass with input derivatives $v_\phi$ for $z_s$, $1$ for $s$, and $0$ for the fixed inputs ($s'$, $X_t$, and $\tilde{X}_{t+1}$) to obtain $u_\phi$ and $\mathrm{d}u_\phi/\mathrm{d}s$ simultaneously.
Because stop-gradient detaches the derivative term, backpropagation differentiates only $u_\phi$, requiring two forward passes and no second-order gradients.

\paragraph{Forward mode through the spherical operators.}
The MLPs, instance normalization, FiLM, and positional embedding support forward mode natively.
The tangent of $s$ enters every block through FiLM, which propagates
\begin{equation}
	\dot{h}' = (1 + a) \odot \dot{h} + \dot{a} \odot h + \dot{b},
	\label{eq:supp_film_jvp}
\end{equation}
where $\dot{a}$ and $\dot{b}$ are projected from $\dot{c} = \mathrm{d}e_s(s) / \mathrm{d}s$ through the block's FiLM linear layer.
The optimized kernels for DISCO and the compiled complex contraction in spectral convolution lack native forward-mode rules.
Because both spherical operators are linear in the activation, their Jacobians coincide with the operators themselves,
\begin{equation}
	(h, \dot{h}) \mapsto (\mathcal{K} h,\, \mathcal{K} \dot{h}), \qquad \mathcal{K} \in \{\mathcal{K}_{\mathrm{G}}, \mathcal{K}_{\mathrm{L}}\}.
	\label{eq:supp_linear_jvp}
\end{equation}
We apply each bias-free DISCO convolution directly to the tangent using the same sparse kernels as the primal pass.
In the spectral convolution, the forward and inverse SHTs propagate tangents natively via FFTs and matrix products, while the complex contraction with $W_\ell$ switches to an equivalent einsum operation whenever tangents are present.
Bilinear upsampling in the decoder propagates tangents identically as sequential matrix products along latitude and longitude.

\subsection{Noise source}
\label{sec:supp_noise}

All source statistics are computed once from the forecast residuals of $F_\theta$ across about 15,000 training days (1979--2019).
We define the standardized one-step residual as
\begin{equation}
	r_{t+1} = \big(X_{t+1} - F_\theta(X_t)\big) \oslash \sigma,
	\label{eq:residual}
\end{equation}
where $\sigma \in \mathbb{R}^C$ is the channelwise standard deviation of $X_{t+1} - F_\theta(X_t)$ computed with the latitude weights of Section~\ref{sec:supp_metrics} over valid grid points, and $\oslash$ denotes channelwise division.
Algorithm~\ref{alg:source} outlines the complete sampling procedure.

\paragraph{Spectral covariance.}
Each daily residual field is expanded in spherical harmonics up to degree $\ell = 120$, the band limit of the $121 \times 240$ grid, to estimate $V_\ell$.
We obtain its Cholesky factor $L_\ell$ after adding diagonal jitter equal to $10^{-6}$ times the mean diagonal entry of $V_\ell$.
Because the spatial field is real, we sample only orders $m \ge 0$.
Specifically, $\xi_\ell^0$ is drawn from a real standard normal distribution, while for $m > 0$ the real and imaginary parts of $\xi_\ell^m$ are independent draws from $\mathcal{N}(0, \tfrac{1}{2} I_C)$, yielding $\mathbb{E}\big[\hat{\varepsilon}_\ell^m (\hat{\varepsilon}_\ell^m)^{\mathrm{H}}\big] = L_\ell L_\ell^\top = V_\ell$ at every order.

\begin{algorithm}[t]
	\caption{Sampling the source of one ensemble member over a $T$-day rollout. Lines 2 to 5 apply to every degree $\ell$ and zonal order $m$.}
	\label{alg:source}
	\begin{algorithmic}[1]
		\Require Cholesky factors $L_\ell$, variance map $w$, static shares $\beta_\ell$, AR(1) coefficients $\rho_\ell$
		\State Draw $\xi^{\mathrm{sh}}, \eta_{t_0} \sim \mathcal{N}(0, I_C)$
		\For{$t = t_0 + 1, \dots, t_0 + T$}
			\State $\eta_t \gets \rho_\ell\, \eta_{t-1} + \sqrt{1 - \rho_\ell^2}\, \zeta_t$, \; $\zeta_t \sim \mathcal{N}(0, I_C)$
			\State $\xi_t \gets \sqrt{\beta_\ell}\, \xi^{\mathrm{sh}} + \sqrt{1 - \beta_\ell}\, \eta_t$
			\State $\hat{\varepsilon}_{t,\ell}^{\,m} \gets L_\ell\, \xi_t$
			\State $\varepsilon_t \gets \sqrt{w} \odot \mathrm{SHT}^{-1} \hat{\varepsilon}_t$
		\EndFor
		\State \Return $\varepsilon_{t_0+1}, \dots, \varepsilon_{t_0+T}$
	\end{algorithmic}
\end{algorithm}

\paragraph{Geographic modulation.}
The variance map $w_c = \overline{r_c^2} \big/ \langle \overline{r_c^2} \rangle$ normalizes the time-averaged squared residual of channel $c$ at each grid point by its latitude-weighted global mean.
We keep $w_c$ at full grid resolution without smoothing, and set it to zero outside the valid ocean and land domains for sea surface temperature and soil moisture.
Scaling the inverse SHT output by $\sqrt{w_c}$ matches the empirical residual variance at every grid point while preserving the spectral cross-channel structure.

\paragraph{Temporal persistence.}
The static share and AR(1) coefficient in Eq.~\eqref{eq:ar1} decay with spherical harmonic degree as
\begin{equation}
	\beta_\ell = \frac{1}{1 + (\ell / \ell_0)^2}, \qquad \rho_\ell = \frac{1}{1 + \ell / \ell_1},
	\label{eq:supp_persistence}
\end{equation}
with $\ell_0 = 5$ and $\ell_1 = 8$.
Under this parameterization, each member keeps most of its planetary-scale noise fixed over the rollout while intermediate scales evolve smoothly from day to day and small-scale perturbations refresh daily.
Consequently, large-scale perturbations build up coherently and accumulate much larger multi-week variance than independent daily draws, without inflating small-scale noise.

We choose these empirical closed forms based on the autocorrelation of one-step residuals.
Fitting $\beta_\ell + (1 - \beta_\ell)\, \rho_\ell^{\,n}$ to the pooled spectral autocorrelations at lags of 1 to 7 days shows strong persistence at planetary wavelengths that tapers smoothly toward synoptic scales.
Summarizing this scale dependence with two cutoff degrees avoids per-degree estimation noise, and setting $\ell_0$ and $\ell_1$ empirically to give stronger persistence than the raw one-step fit compensates for how multi-week rollouts damp inter-member spread.
The Markov chain operates on the standard normal coefficients $\xi_\ell^m$ before applying $L_\ell$, independently across degrees, orders, and channels.
For sea surface temperature and the two soil moisture layers, we redraw $\xi$ independently each day ($\beta_\ell = \rho_\ell = 0$) to prevent their over-dispersion, because these slowly varying surface states already integrate daily residuals over time.
Because the single-step marginal stays $\mathcal{N}(0, I_C)$, the persistent chain plugs directly into the Stage~2 generator without shifting its input distribution, and each ensemble member runs an independent chain in Stage~3 and at inference.

\subsection{Computational cost}
\label{sec:supp_cost}

Here we describe how the computational figures in Section~\ref{sec:efficiency} are measured.

\paragraph{Parameters and FLOPs.}
Parameter counts come directly from the instantiated networks, and FLOPs are computed analytically for a single forward pass at batch size one, counting each multiply--add as two operations.
For DISCO convolutions, we count the nonzero entries of the sparse kernels plus channel mixing, and we include the spherical harmonic transforms.
VistaCast totals 47.0\,M parameters (30.6\,M in $F_\theta$ and 16.4\,M in $G_\phi$) and requires 0.31\,TFLOPs per member-day (200\,GFLOPs for $F_\theta$ and 112\,GFLOPs for $G_\phi$).
We compute the FuXi-S2S figures identically from its official ONNX release, excluding the training-only posterior encoder.

\paragraph{Training and inference time.}
The complete three-stage training of VistaCast takes roughly 140 A800 GPU hours.
FuXi-S2S reports training on eight A100 GPUs without specifying wall-clock duration~\citep{chenMachineLearningModel2024}.
VistaCast runs a 51-member ensemble (including source sampling and the control member) as a single batch in 106\,s, or 2.1\,s per member.
The official ONNX release of FuXi-S2S uses a fixed batch size of one and takes 8.05\,s per member on the same hardware, consistent with the speed reported by its authors~\citep{chenMachineLearningModel2024}.

\paragraph{Why one-step generation matters.}
Unrolling $B$ initial states over $H$ days with $M$ members builds a computation graph of $B \times M \times H$ sequential transitions.
With activation checkpointing on every processor block, an 80\,GB GPU accommodates about 144 transitions, comfortably fitting $B = 1$, $M = 3$, and $H = 42$ (126 transitions).
Iterative diffusion or flow samplers that need tens of network evaluations per step would multiply this graph by the sampling step count, exceeding GPU memory and slowing ensemble inference by nearly an order of magnitude.

\section{Experimental Details}
\label{sec:supp_experiments}

\subsection{Data}
\label{sec:supp_data}

We train and evaluate on ERA5~\citep{hersbachERA5GlobalReanalysis2020}, the ECMWF global reanalysis that combines a numerical weather model with global observations into a physically consistent historical record.
{\urlstyle{same}\Urlmuskip=0mu\relax ERA5 can be downloaded from the Copernicus Climate Data Store (\url{https://cds.climate.copernicus.eu}), and it is also distributed in \mbox{WeatherBench~2}~\citep{raspWeatherBench2Benchmark2024a} (\url{https://console.cloud.google.com/storage/browser/weatherbench2/datasets/era5}).
We take the daily-mean $1.5^\circ$ ERA5 dataset on a $121 \times 240$ equiangular grid including both poles.}
Table~\ref{tab:supp_variables} lists all input and output variables.
We standardize each channel using its training-set mean and standard deviation.
Sea surface temperature and soil moisture are defined only over ocean and land points, respectively, and are masked accordingly in all inputs, losses, and metrics.
Following ChaosBench~\citep{nathanielChaosBenchMultiChannelPhysicsBased2024}, we split the data into 1979--2019 for training, 2020--2021 for validation, and 2022 for testing.

\subsection{Baselines}
\label{sec:supp_protocol}

We benchmark VistaCast against three operational NWP systems, two medium-range AI models, three AI S2S models, and climatology.
\begin{itemize}
	\item \textbf{ECMWF} is the extended-range ensemble from the European Centre for Medium-Range Weather Forecasts, based on its ocean-coupled Integrated Forecasting System (IFS), cycle 41r1~\citep{molteniECMWFEnsemblePrediction1996, vitartEvolutionECMWFSubseasonal2014}.
	\item \textbf{NCEP} is the Climate Forecast System version 2 (CFSv2) from the National Centers for Environmental Prediction, coupling atmosphere, ocean, land, and sea-ice components~\citep{sahaNCEPClimateForecast2014}.
	\item \textbf{UKMO} is the UK Met Office Global Seasonal forecast system version 6 (GloSea6), built on the coupled GC3.2 atmosphere--ocean model~\citep{kettleboroughGlobalSeasonalForecasting2025}.
	\item \textbf{Pangu-Weather} is a 3D Earth-specific Transformer for $0.25^\circ$ medium-range forecasting that chains separate 1-, 3-, 6-, and 24-hour models through hierarchical temporal aggregation~\citep{biAccurateMediumrangeGlobal2023}.
	\item \textbf{GraphCast} is a $0.25^\circ$ graph neural network for medium-range forecasting that passes messages on a multiscale icosahedral mesh and predicts 6-hour steps from two preceding states~\citep{lamLearningSkillfulMediumrange2023}.
	\item \textbf{FuXi-S2S} is a $1.5^\circ$ Transformer-based S2S model forecasting up to 42 days ahead by perturbing hidden features with state-dependent Gaussian noise~\citep{chenMachineLearningModel2024}.
	\item \textbf{CirT} is an S2S Transformer that tokenizes the globe into latitude rings and applies Fourier transforms inside self-attention to respect longitudinal periodicity~\citep{liuCirTGlobalSubseasonaltoSeasonal2025a}.
	\item \textbf{TelePiT} is a physics-informed S2S Transformer that uses spherical harmonic embeddings, neural ordinary differential equations for multiscale dynamics, and teleconnection-aware self-attention~\citep{lyuPhysicsInformedTeleconnectionAwareTransformer2025}.
	\item \textbf{Climatology} is the 1981--2010 ERA5 calendar-day mean, providing a reference forecast with no initial-state information.
\end{itemize}

\begin{table}[t]
	\caption{
		Variables used by VistaCast.
		Upper-air variables are given at 50, 100, 200, 250, 300, 500, 700, 850, 925, and 1000\,hPa.
	}
	\label{tab:supp_variables}
	\centering
	\small
	\setlength{\tabcolsep}{4pt}
	\begin{tabular}{@{}llll@{}}
		\toprule
		Type & Variable & Abbrev. & Role \\
		\midrule
		Upper-air & Geopotential & Z & In/out \\
		& Specific humidity & Q & In/out \\
		& Temperature & T & In/out \\
		& Zonal wind & U & In/out \\
		& Meridional wind & V & In/out \\
		\midrule
		Single-level & 10\,m zonal wind & U10 & In/out \\
		& 10\,m meridional wind & V10 & In/out \\
		& 2\,m temperature & T2M & In/out \\
		& Mean sea-level pressure & MSLP & In/out \\
		& Total column water vapor & TCWV & In/out \\
		& Outgoing longwave radiation & OLR & In/out \\
		& Sea surface temperature & SST & In/out \\
		& Soil moisture, 7--28\,cm & SWVL2 & In/out \\
		& Soil moisture, 28--100\,cm & SWVL3 & In/out \\
		\midrule
		Static & Land--sea mask & LSM & In \\
		\bottomrule
	\end{tabular}
\end{table}

Models have different temporal targets.
The NWP systems, Pangu-Weather, and GraphCast predict instantaneous fields, FuXi-S2S and VistaCast predict daily means, while CirT and TelePiT directly predict biweekly means for weeks 3--4 and 5--6.
We target daily means in VistaCast for three reasons.
(1) Beyond two weeks, setting instantaneous values as targets introduces unnecessary noise.
At S2S scales, sub-daily fluctuations are no longer predictable, and forecast skill comes from slower modes that can be more naturally represented with daily means.
(2) Compared with weekly or biweekly objectives, daily trajectories preserve day-to-day evolution and can be averaged directly into the biweekly means.
(3) Daily means avoid the mixed diurnal phases introduced when a single 00\,UTC snapshot is scored globally.
Daily means are also standard in recent AI S2S models such as FuXi-S2S and TianQuan-S2S~\citep{chenMachineLearningModel2024, liTIANQUANS2SSUBSEASONALTOSEASONALGLOBAL2026}.

We obtain the 2022 scores of the NWP systems, Pangu-Weather, and GraphCast directly from ChaosBench~\citep{nathanielChaosBenchMultiChannelPhysicsBased2024}.
For FuXi-S2S, we run its official release with 51 members on the same 2022 initializations as VistaCast and evaluate both with the same pipeline.
Every model is verified against matching ERA5 targets.
The biweekly comparison in Section~\ref{sec:supp_biweekly} follows the same convention.

\subsection{Evaluation metrics}
\label{sec:supp_metrics}

We compute all metrics separately at each lead time.
Let $x_{n,i,j}$ denote the ERA5 ground truth at initialization $n \in \{1, \dots, N\}$, latitude $i \in \{1, \dots, H\}$, and longitude $j \in \{1, \dots, W\}$, and let $\hat{x}^{(m)}_{n,i,j}$ be the forecast of member $m \in \{1, \dots, M\}$ with ensemble mean $\bar{x}_{n,i,j}$.
All spatial averages use the latitude weights $a_i = \cos\varphi_i \big/ \big(\frac{1}{H}\sum_{i'} \cos\varphi_{i'}\big)$, where $\varphi_i$ is the latitude of row $i$.
All sums exclude masked grid points.

\paragraph{RMSE and ACC.}
The ensemble-mean RMSE is
\begin{equation}
	\mathrm{RMSE} = \sqrt{\frac{1}{NHW} \sum_{n,i,j} a_i \big(\bar{x}_{n,i,j} - x_{n,i,j}\big)^2 },
	\label{eq:supp_rmse}
\end{equation}
where $\bar{x}$ is replaced by the single forecast for deterministic models.
Deterministic evaluations of VistaCast use the control member $F_\theta$.
The anomaly correlation coefficient is
\begin{equation}
	\mathrm{ACC} = \frac{\sum_{n,i,j} a_i\, \bar{x}'_{n,i,j}\, x'_{n,i,j}}{\sqrt{\sum_{n,i,j} a_i\, \bar{x}'^{\,2}_{n,i,j} \sum_{n,i,j} a_i\, x'^{\,2}_{n,i,j}}},
	\label{eq:supp_acc}
\end{equation}
where anomalies $\bar{x}'$ and $x'$ are formed by subtracting the 1981--2010 calendar-day climatology of the valid date.

\paragraph{CRPS, spread, and SSR.}
As a strictly proper scoring rule that generalizes mean absolute error to predictive distributions~\citep{gneitingStrictlyProperScoring2007}, CRPS is estimated as
\begin{equation}
	\begin{aligned}
		\mathrm{CRPS} = \frac{1}{NHW} \sum_{n,i,j} a_i \Bigg[ & \frac{1}{M} \sum_{m} \big|\hat{x}^{(m)}_{n,i,j} - x_{n,i,j}\big| \\
		& - \frac{1}{2M^2} \sum_{m, m'} \big|\hat{x}^{(m)}_{n,i,j} - \hat{x}^{(m')}_{n,i,j}\big| \Bigg].
	\end{aligned}
	\label{eq:supp_crps}
\end{equation}
The ensemble spread is
\begin{equation}
	\mathrm{Spread} = \sqrt{\frac{1}{NHW} \sum_{n,i,j} a_i\, \frac{M+1}{M}\, s^2_{n,i,j}},
	\label{eq:supp_spread}
\end{equation}
where $s^2_{n,i,j}$ is the unbiased ensemble variance and $(M+1)/M$ corrects for finite ensemble size.
The spread--skill ratio $\mathrm{SSR} = \mathrm{Spread} / \mathrm{RMSE}$ measures calibration, with values below 1 indicating under-dispersion and values above 1 indicating over-dispersion.

\begin{figure*}[t]
	\centering
	\includegraphics[width=\linewidth]{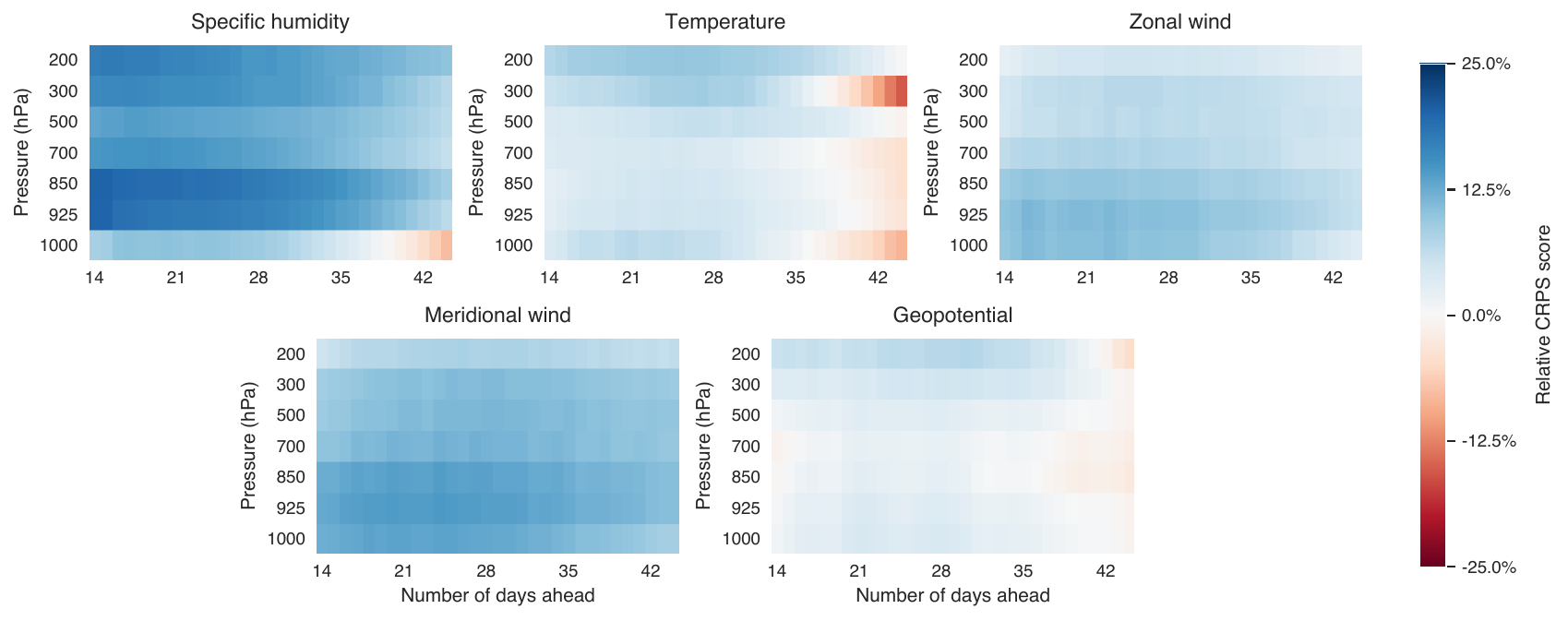}
	\caption{
		Relative CRPS of VistaCast against the ECMWF ensemble on the 2022 test year for five upper-air variables at seven pressure levels and leads from 14 to 44 days.
		Each cell shows $100\,(1 - \mathrm{CRPS}_{\mathrm{VistaCast}} / \mathrm{CRPS}_{\mathrm{ECMWF}})$.
	}
	\label{fig:supp_levels}
\end{figure*}

\paragraph{CRPSS.}
The continuous ranked probability skill score normalizes CRPS against a climatological reference,
\begin{equation}
	\mathrm{CRPSS} = 1 - \frac{\mathrm{CRPS}}{\mathrm{CRPS}_{\mathrm{clim}}},
	\label{eq:supp_crpss}
\end{equation}
where positive scores indicate skill above climatology.
Following ChaosBench~\citep{nathanielChaosBenchMultiChannelPhysicsBased2024}, the reference at each grid point is a Gaussian distribution parameterized by the 1981--2010 calendar-day mean and standard deviation, and $\mathrm{CRPS}_{\mathrm{clim}}$ is the latitude-weighted average of its closed-form CRPS.

\paragraph{Physical consistency.}
Pointwise error metrics alone can reward blurred forecasts that violate atmospheric physics, so we evaluate three structural diagnostics on individual ensemble members.
First, in the mid-latitudes (20--70$^\circ$ in both hemispheres), large-scale horizontal winds are governed primarily by geostrophic balance between the horizontal pressure-gradient force and the Coriolis force from Earth's rotation.
This balance links the horizontal wind vector $\mathbf{V}$ to the spatial gradient of geopotential $\Phi$ via the geostrophic wind $\mathbf{V}_g = f^{-1}\, \mathbf{k} \times \nabla \Phi$, where $f$ is the Coriolis parameter and $\nabla \Phi$ is computed by centered differences.
We measure departures from this balance using the ageostrophic wind fraction $\sqrt{\langle |\mathbf{V} - \mathbf{V}_g|^2 \rangle / \langle |\mathbf{V}_g|^2 \rangle}$, where $\langle \cdot \rangle$ denotes an area-weighted mean over the mid-latitudes.
Because real weather systems contain genuine ageostrophic motion, a physically consistent forecast should stay close to the ERA5 reference value across lead times, with higher values indicating inconsistent wind--pressure coupling.
Vertically, large-scale atmospheric columns generally obey hydrostatic balance between the upward pressure-gradient force and gravity, tying the geopotential thickness between two pressure levels to the layer's mean virtual temperature $T_v = T(1 + 0.608\, q)$.
We quantify hydrostatic consistency by the root mean square of the thickness residual,
\begin{equation}
	\delta = \frac{\Phi_{\mathrm{top}} - \Phi_{\mathrm{bottom}}}{g} - \frac{R_d}{g} \int_{p_{\mathrm{top}}}^{p_{\mathrm{bottom}}} T_v\, \mathrm{d}\ln p,
	\label{eq:supp_hydrostatic}
\end{equation}
where $R_d$ is the dry-air gas constant.
We integrate by the trapezoidal rule across the 850--700--500\,hPa and 500--300--250\,hPa layers and exclude grid points where a layer falls below the 2022 mean ERA5 surface pressure.
Smaller residuals close to the ERA5 baseline indicate coherent vertical coupling among predicted temperature, humidity, and geopotential.
Third, the spatial power spectrum decomposes field variance across horizontal scales indexed by spherical harmonic degree $\ell$, from planetary waves at low $\ell$ to fine-scale weather structures at high $\ell$,
\begin{equation}
	P(\ell) = \frac{1}{4\pi} \sum_{m=-\ell}^{\ell} |\hat{f}_\ell^m|^2.
	\label{eq:supp_spectrum}
\end{equation}
A realistic forecast matches the ERA5 spectrum across all wavenumbers, avoiding the high-wavenumber energy drop caused by blurring as well as excess power from grid-scale noise.

\subsection{Training and inference}
\label{sec:supp_training}

All three stages use AdamW~\citep{loshchilovDecoupledWeightDecay2019} with a weight decay of 0.01 and bfloat16 mixed precision.
Table~\ref{tab:supp_curriculum} summarizes the curriculum schedule for each stage.

\paragraph{Stage 1.}
Unrolling $F_\theta$ over $K$ daily steps, we minimize
\begin{equation}
	\mathcal{L}_{\mathrm{det}} = \frac{1}{\sum_k \lambda_k} \sum_{k=1}^{K} \lambda_k \, \big\| F_\theta^{(k)}(X_{t_0}) - X_{t_0+k} \big\|_A^2,
	\label{eq:det_loss}
\end{equation}
where $F_\theta^{(k)}$ denotes $k$ autoregressive steps of $F_\theta$ and $\|\cdot\|_A^2$ is the latitude-weighted mean squared norm.
The lead weight $\lambda_k$ stays at 1 during the first week, ramps linearly to 2 by day~14, and remains at 2 thereafter.
We also weight training samples linearly from 0.9 in January 1979 to 1.1 in December 2019 to place slightly more emphasis on recent climate conditions.
The horizon $K$ expands from 1 to 42 days across 15 phases.
Epoch-based phases apply cosine decay to 0.1 of the peak learning rate, while fixed-step phases hold the learning rate constant, with AdamW momentum $\beta = (0.9, 0.95)$.

\begin{figure*}[t]
	\centering
	\includegraphics[width=\linewidth]{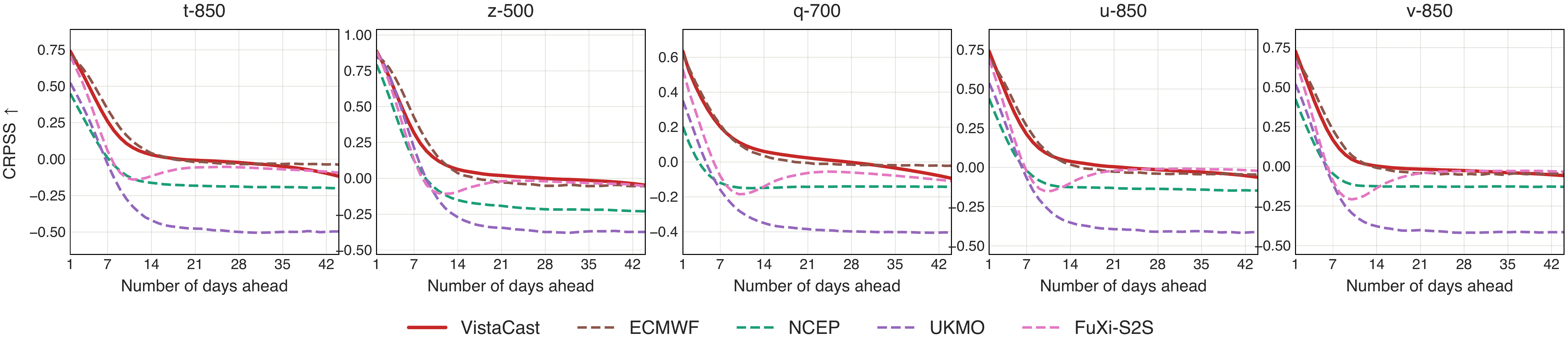}
	\caption{
		CRPSS on the 2022 test year against lead time for five key variables (higher is better), where positive values indicate skill beyond the climatological reference forecast.
	}
	\label{fig:supp_crpss}
\end{figure*}

\paragraph{Stage 2.}
We first train $G_\phi$ to model the residual, then continue on transitions from deterministic rollouts of depth $k \in \{1, 2, 3, 4\}$.
Precomputing these rollout states offline with the frozen $F_\theta$ keeps Stage~2 at the cost of standard single-step flow training.
Both phases use $\beta = (0.9, 0.98)$ with a 10\% linear warm-up and cosine decay to 0.1 of the peak learning rate.
The output layer of the lead embedding $e_\kappa$ remains frozen at zero during the first phase, which only observes single-day transitions, and trains at twice the base learning rate in the second phase.

\begin{table}[t]
	\caption{
		Training curriculum by stage.
		Lead is the rollout horizon in days for Stages~1 and~3 and the depth of the deterministic rollout supplying the input for Stage~2.
	}
	\label{tab:supp_curriculum}
	\centering
	\footnotesize
	\setlength{\tabcolsep}{3pt}
	\newcommand{\curriculumhead}[2]{\multicolumn{4}{c}{\textbf{Stage #1}} \\ \cmidrule{1-4} \textbf{Lead} & \textbf{#2} & \textbf{Batch} & \textbf{LR} \\ \midrule}
	\newsavebox{\curriculumright}
	\newsavebox{\curriculumleft}
	\newlength{\curriculumgap}
	\sbox{\curriculumright}{\begin{tabular*}{0.44\columnwidth}[t]{@{\extracolsep{\fill}}rrrr@{}}
		\curriculumhead{2}{Steps}
		\textbf{1} & 25k & 25 & 4e-4 \\
		\textbf{1--4} & 12k & 25 & 5e-5 \\
		\midrule
		\curriculumhead{3}{Steps}
		\textbf{2} & 300 & 16 & 2e-5 \\
		\textbf{3} & 300 & 12 & 2e-5 \\
		\textbf{7} & 300 & 5 & 1.6e-5 \\
		\textbf{10} & 200 & 4 & 1.4e-5 \\
		\textbf{12} & 200 & 3 & 1.2e-5 \\
		\textbf{14} & 200 & 3 & 1e-5 \\
		\textbf{18} & 200 & 2 & 1e-5 \\
		\textbf{21} & 200 & 2 & 1e-5 \\
		\textbf{24} & 200 & 1 & 1e-5 \\
		\textbf{28} & 200 & 1 & 1e-5 \\
		\textbf{32} & 200 & 1 & 6e-6 \\
		\textbf{36} & 200 & 1 & 6e-6 \\
		\textbf{42} & 1200 & 1 & 4e-6
	\end{tabular*}}
	\newcommand{\curriculumstageone}{\begin{tabular*}{\dimexpr 0.56\columnwidth - 14pt\relax}[t]{@{\extracolsep{\fill}}rrrr@{}}
		\curriculumhead{1}{Epochs/steps}
		\textbf{1} & 100\,epochs & 42 & 5e-3 \\[\curriculumgap]
		\textbf{2} & 10\,epochs & 24 & 6e-4 \\[\curriculumgap]
		\textbf{3} & 3\,epochs & 18 & 5e-5 \\[\curriculumgap]
		\textbf{5} & 1\,epoch & 12 & 3e-5 \\[\curriculumgap]
		\textbf{7} & 2\,epochs & 8 & 2e-5 \\[\curriculumgap]
		\textbf{10} & 1\,epoch & 6 & 2e-5 \\[\curriculumgap]
		\textbf{12} & 1\,epoch & 5 & 2e-5 \\[\curriculumgap]
		\textbf{14} & 600\,steps & 4 & 1e-5 \\[\curriculumgap]
		\textbf{16} & 600\,steps & 4 & 6e-6 \\[\curriculumgap]
		\textbf{18} & 600\,steps & 3 & 4e-6 \\[\curriculumgap]
		\textbf{21} & 600\,steps & 3 & 2e-6 \\[\curriculumgap]
		\textbf{24} & 600\,steps & 2 & 2e-6 \\[\curriculumgap]
		\textbf{32} & 600\,steps & 2 & 1e-6 \\[\curriculumgap]
		\textbf{36} & 600\,steps & 1 & 1e-6 \\[\curriculumgap]
		\textbf{42} & 4200\,steps & 1 & 1e-6
	\end{tabular*}}
	\setlength{\curriculumgap}{0pt}
	\sbox{\curriculumleft}{\curriculumstageone}
	\setlength{\curriculumgap}{\dimexpr(\dp\curriculumright - \dp\curriculumleft) / 14\relax}
	\sbox{\curriculumleft}{\curriculumstageone}
	\begin{tabular}{@{}l@{\hspace{14pt}}l@{}}
		\toprule
		\usebox{\curriculumleft} & \usebox{\curriculumright} \\
		\bottomrule
	\end{tabular}
\end{table}

\paragraph{Stage 3.}
From each initial state, we unroll three ensemble members with independent source chains and backpropagate the afCRPS loss (Eq.~\eqref{eq:afcrps}) through the full trajectory, including $F_\theta$.
This teaches $G_\phi$ how daily residuals evolve through subsequent deterministic steps to maintain spread weeks ahead.
Setting $\alpha = 1$ in afCRPS yields the fair CRPS, an unbiased estimator for finite ensembles that is critical when $M = 3$, and lowering it slightly to $\alpha = 0.95$ prevents degenerate spread configurations~\citep{langAIFSCRPSEnsembleForecasting2026}.
The rollout horizon grows across 13 phases using the optimizer and learning-rate schedule of Stage~2.
Starting from $H = 10$, we jointly finetune $F_\theta$ at 0.2 times the base learning rate with gradient norm clipping at 2.0, while training the lead embedding at $10\times$ the base learning rate ($20\times$ at $H = 42$).
The control member retains the Stage~1 weights of $F_\theta$.

\paragraph{Inference.}
We advance 50 perturbed members via Eq.~\eqref{eq:transition} using independent source chains and one generator evaluation per day, alongside one control member driven solely by the Stage~1 $F_\theta$.
Forecasts are initialized daily throughout 2022, running all 51 members of an initialization in a single batch.
Before adding each daily residual, the inflation in Eq.~\eqref{eq:inflation} rescales upper-air member deviations $\delta^{(m)} = \tilde{X}^{(m)}_{t+1} - \overline{\tilde{X}}_{t+1}$ in spectral space,
\begin{equation}
	\hat{\delta}^{(m)}_{\ell} \leftarrow g_k\, \hat{\delta}^{(m)}_{\ell} \quad \text{for } \ell \le 10,
	\label{eq:supp_inflation}
\end{equation}
where $\hat{\delta}^{(m)}_\ell$ holds the coefficients of degree $\ell$ and $g_k = 1 + 0.05\, k / 44$ increases linearly with lead day $k$.
Because member deviations sum to zero, this leaves the ensemble mean untouched and scales only planetary-scale spread.
This step counteracts the residual spread damping introduced by $F_\theta$, which predicts conditional means.
Surface channels keep $g_k = 1$, and inflation is applied only at inference without altering training.

\section{Additional Results}
\label{sec:supp_results}

\subsection{Comparison with biweekly models}
\label{sec:supp_biweekly}

We also compare VistaCast against recent direct biweekly-mean S2S models, namely CirT~\citep{liuCirTGlobalSubseasonaltoSeasonal2025a} and TelePiT~\citep{lyuPhysicsInformedTeleconnectionAwareTransformer2025}.
To match their targets, we average VistaCast's daily forecasts over leads 15--28 (weeks 3--4) and 29--42 (weeks 5--6) and verify them against the corresponding ERA5 means.
We retrain CirT from scratch using its official code and split (1979--2016 training, 2017 validation) on daily-mean ERA5, and evaluate TelePiT's released checkpoint on its ChaosBench targets.
We omit TelePiT's T2M score because its normalization issue inflates the RMSE to about 7\,K, making it incomparable with the other models.
All models are initialized and verified over 2022.
As shown in Table~\ref{tab:supp_biweekly}, VistaCast's ensemble mean attains the lowest RMSE across all seven variables in weeks 3--4, and on Z500, T500, and T2M in weeks 5--6.
Because CirT and TelePiT predict biweekly averages directly, they avoid autoregressive error accumulation and show little RMSE growth between the two forecast windows.
At the same time, VistaCast resolves daily weather evolution and generates sharp probabilistic ensembles with calibrated uncertainty, while these deterministic baselines output blurred biweekly means and cannot be evaluated on probabilistic metrics such as CRPS.

\par\medskip\noindent\begin{minipage}{\columnwidth}
	\captionof{table}{
		Biweekly-mean RMSE on the 2022 test year (lower is better), with the lowest error in each window in bold.
	}
	\label{tab:supp_biweekly}
	\centering
	\footnotesize
	\setlength{\tabcolsep}{2pt}
	\setlength{\aboverulesep}{0pt}
	\setlength{\belowrulesep}{0pt}
	\renewcommand{\arraystretch}{1.35}
	\begin{tabular*}{\columnwidth}{@{\hspace{6pt}}c@{\hspace{6pt}}|@{\extracolsep{\fill}\hspace{3pt}}l|*{7}{c}@{\hspace{1pt}}}
		\toprule
		& \textbf{Model} & \textbf{Z500} & \textbf{T850} & \textbf{T500} & \textbf{Q700} & \textbf{U850} & \textbf{V850} & \textbf{T2M} \\[-3pt]
		&& \scriptsize (gpm)
		& \scriptsize (K) & \scriptsize (K) & \scriptsize (g/kg)
		& \scriptsize (m/s) & \scriptsize (m/s) & \scriptsize (K) \\
		\midrule
		\multirow{3}{*}{\rotatebox[origin=c]{90}{\bfseries\begin{tabular}{@{}c@{}}Weeks\\3--4\end{tabular}}}
		& \textbf{CirT} & 51.64 & 2.007 & 1.722 & 0.774 & 2.799 & 2.079 & 2.093 \\
		& \textbf{TelePiT} & 51.09 & 2.042 & 1.794 & 0.871 & 3.033 & 2.152 & -- \\
		& \textbf{VistaCast} & \textbf{46.64} & \textbf{1.915} & \textbf{1.549} & \textbf{0.698} & \textbf{2.679} & \textbf{2.015} & \textbf{1.736} \\
		\midrule
		\multirow{3}{*}{\rotatebox[origin=c]{90}{\bfseries\begin{tabular}{@{}c@{}}Weeks\\5--6\end{tabular}}}
		& \textbf{CirT} & 50.54 & \textbf{2.033} & 1.717 & \textbf{0.779} & \textbf{2.809} & \textbf{2.052} & 2.095 \\
		& \textbf{TelePiT} & 50.44 & 2.064 & 1.805 & 0.874 & 3.069 & 2.148 & -- \\
		& \textbf{VistaCast} & \textbf{49.88} & 2.060 & \textbf{1.630} & 0.810 & 2.910 & 2.170 & \textbf{2.028} \\
		\bottomrule
	\end{tabular*}
\end{minipage}\par\medskip

\subsection{Skill across pressure levels}
\label{sec:supp_levels}

Figure~\ref{fig:supp_levels} extends our CRPS comparison against ECMWF to five upper-air variables across seven pressure levels from 200 to 1000\,hPa over leads 14 to 44.
VistaCast achieves lower CRPS in 93\% of the 1085 evaluated cells, winning all but one cell during weeks 3--4.
After day 36, VistaCast trails ECMWF mainly on temperature and geopotential.

\subsection{Skill relative to climatology}
\label{sec:supp_crpss}

Figure~\ref{fig:supp_crpss} compares the climatology-referenced skill score (CRPSS) across the five key variables.
VistaCast sustains positive skill above climatology longer than every baseline on all five variables, extending to day 27 on Z500.
ECMWF loses skill between days 14 and 19, and the remaining ensemble baselines fall below climatology within the first nine days.

\begin{figure*}[t]
	\centering
	\includegraphics[width=\linewidth]{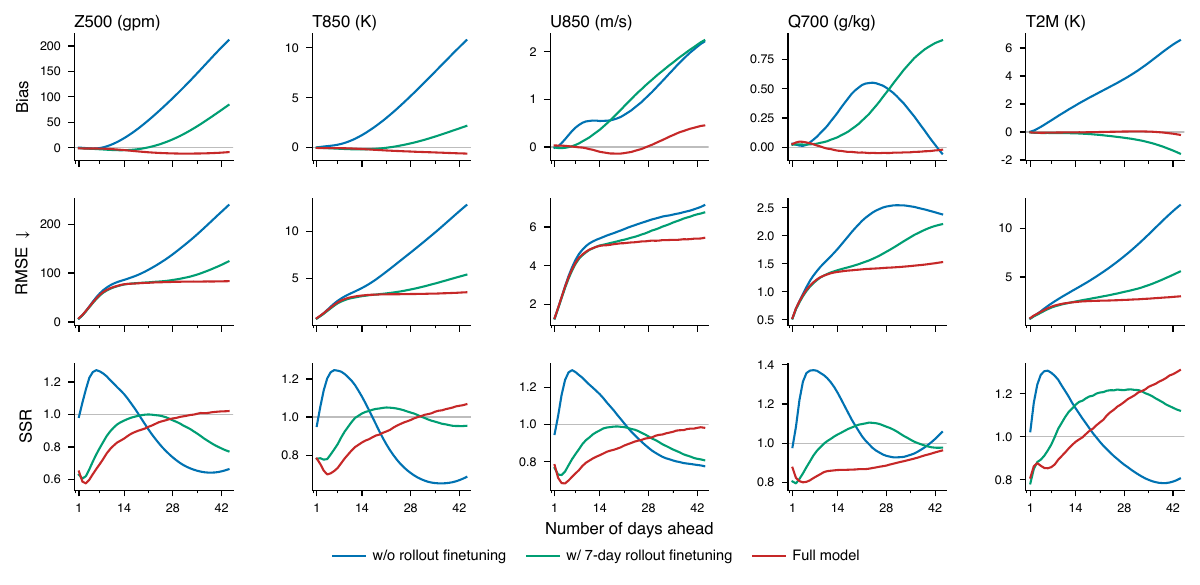}
	\caption{
		Ensemble error of VistaCast across three training stages on the 2022 test year, showing ensemble-mean global bias (top), RMSE (middle), and SSR (bottom).
		The stages correspond to the end of Stage~2 (w/o rollout finetuning), Stage~3 stopped at $H = 7$ before $F_\theta$ joins finetuning (w/ 7-day rollout finetuning), and the full Stage~3 model trained up to $H = 42$ (full model).
		All stages use 25 members with single-step sampling and spectral inflation, initialized every third day in 2022 (107 dates).
		Gray lines mark zero bias and $\mathrm{SSR} = 1$.
	}
	\label{fig:supp_rollout}
\end{figure*}

\subsection{Effect of rollout finetuning}
\label{sec:supp_rollout}

To see how rollout finetuning improves long-lead skill, Figure~\ref{fig:supp_rollout} compares the global-mean bias, RMSE, and SSR across three training stages.

Without rollout finetuning, the ensemble mean drifts steadily from the second week onward, creating large global biases that drive most of the long-lead error growth.
Finetuning over a 7-day horizon delays this drift by a few weeks, yet substantial bias still appears at longer leads.
Extending finetuning to 42 days keeps the bias near zero across all five variables through day 44, allowing the RMSE to level off at long leads while the earlier stages keep diverging.

Mean drift also governs ensemble calibration.
Without rollout finetuning, the ensemble becomes over-dispersive in the first week and then drops sharply into under-dispersion once mean drift inflates forecast error.
Although rollout finetuning slightly reduces raw ensemble spread, suppressing the mean drift keeps errors bounded and guides the SSR smoothly toward one across upper-air variables.
\begin{figure*}[t]
	\centering
	\includegraphics[width=\linewidth]{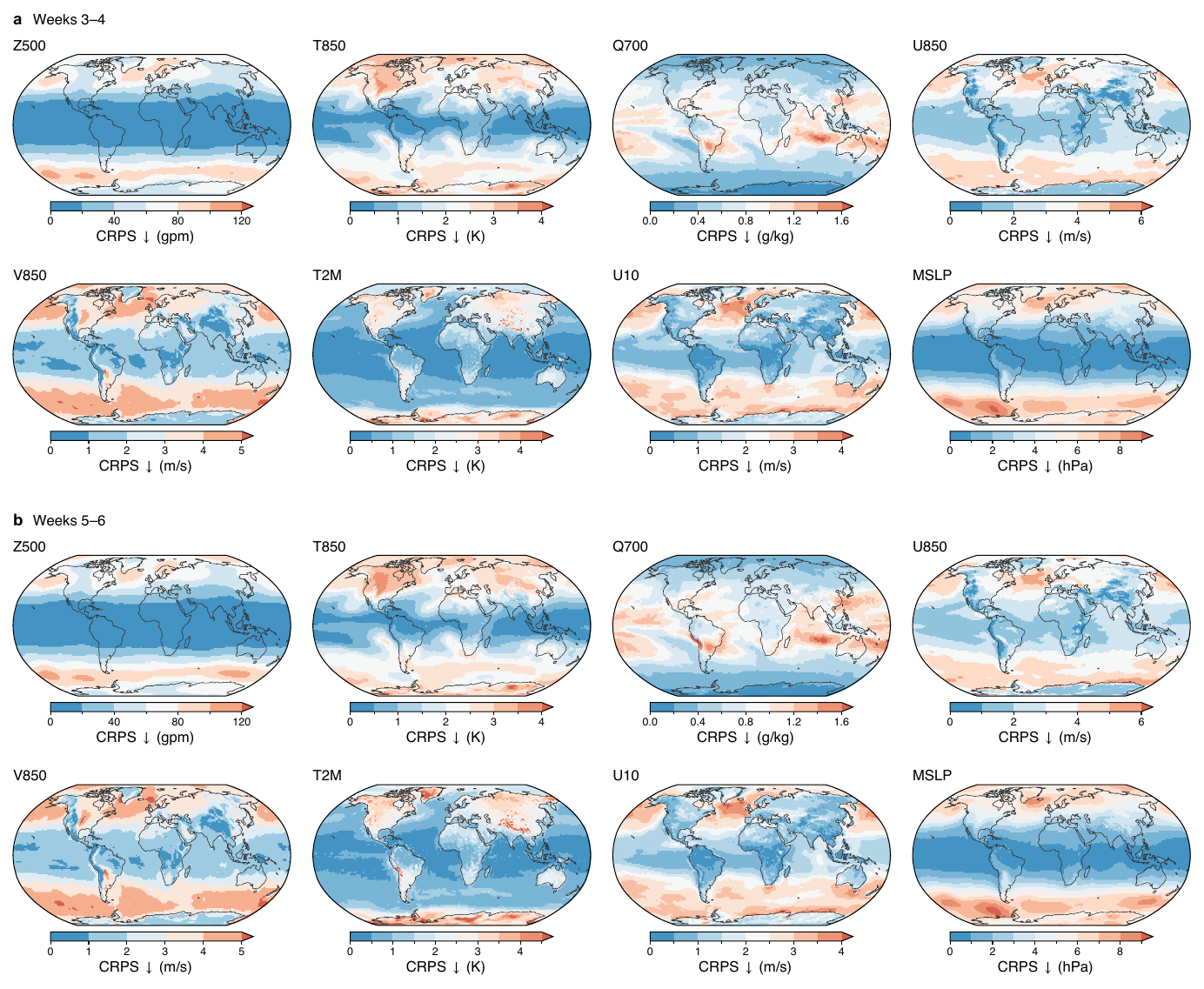}
	\caption{
		Spatial CRPS distribution of VistaCast on the 2022 test year, averaged over leads 15--28 (a, weeks 3--4) and 29--42 (b, weeks 5--6), for eight key variables.
		Each variable shares its color levels between the two windows, and blue indicates a lower CRPS.
	}
	\label{fig:supp_maps}
\end{figure*}

\subsection{Spatial distribution of skill}
\label{sec:supp_maps}

Figure~\ref{fig:supp_maps} maps VistaCast's grid-point CRPS averaged over weeks 3--4 and 5--6.
For dynamical fields such as geopotential, sea-level pressure, and wind, errors follow natural atmospheric variability, staying smallest in the tropics and peaking along extratropical storm tracks, especially over the Southern Ocean.
The 2\,m temperature (T2M) remains accurate over the oceans, where slowly varying sea surface temperatures anchor the air above, while errors concentrate over high-latitude and Northern Hemisphere landmasses.
From weeks 3--4 to weeks 5--6, these spatial structures remain virtually unchanged and global-mean CRPS grows only slightly for most variables, reflecting stable long-lead behavior.

\subsection{Skill in the first two weeks}
\label{sec:supp_short}

\begin{figure*}[t]
	\centering
	\includegraphics[width=\linewidth]{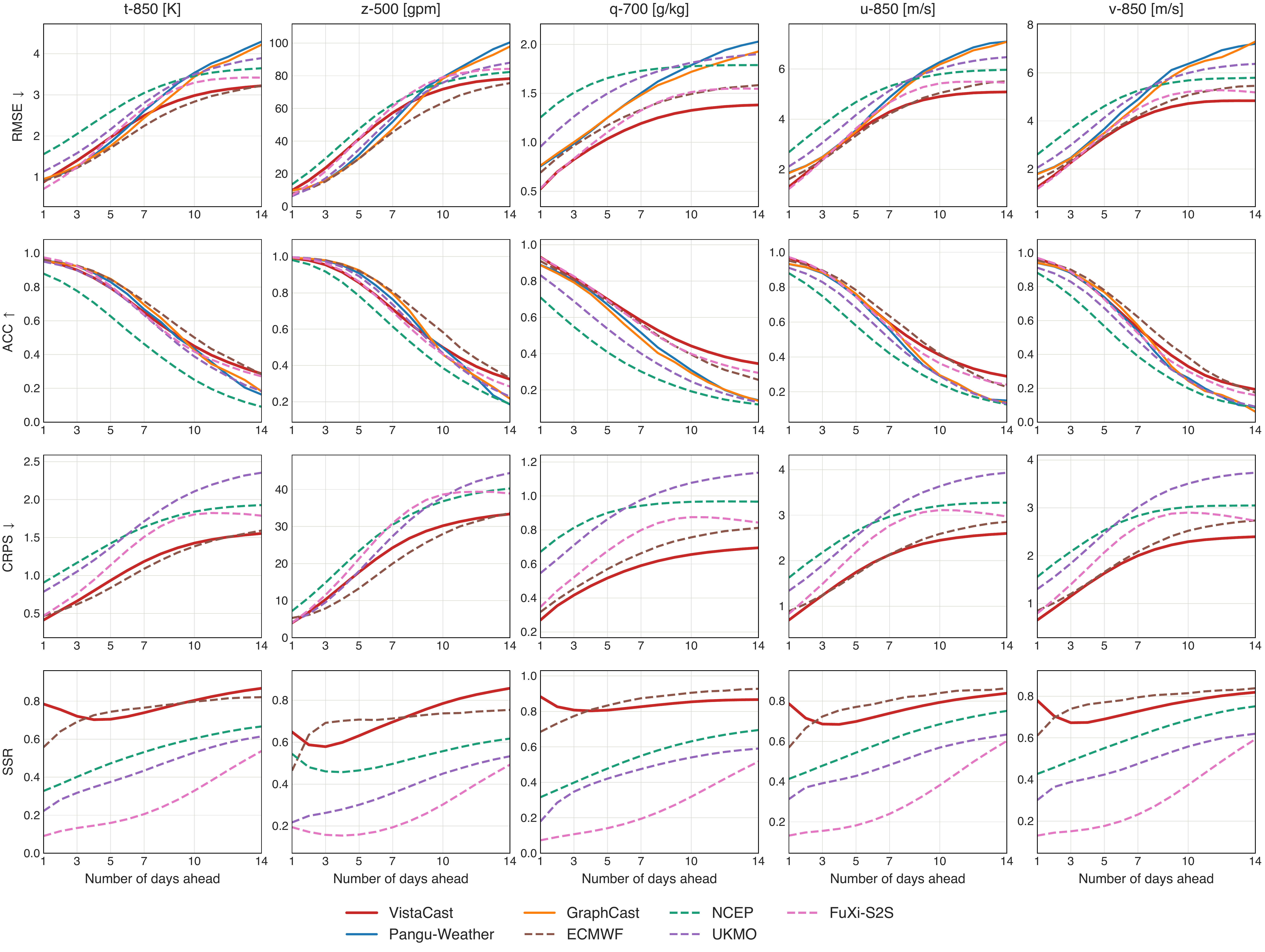}
	\caption{
		Forecast skill on the 2022 test year over the first 14 days for five key variables, showing RMSE, ACC, CRPS, and SSR from top to bottom.
		The climatology is omitted.
	}
	\label{fig:supp_short}
\end{figure*}

Figure~\ref{fig:supp_short} shows forecast skill over the first 14 days.
Despite being finetuned over long 42-day rollouts, VistaCast preserves strong forecast accuracy during the first two weeks.
VistaCast achieves the lowest V850 CRPS across all leads and the lowest U850 CRPS at every lead except days 4 to 6.
Its RMSE is better than that of Pangu-Weather and GraphCast across all evaluated variables from day 10 onward.
Throughout both weeks, VistaCast remains better calibrated than NCEP, UKMO, and FuXi-S2S, which suffer from severe under-dispersion during the first week.

\end{document}